\documentclass[a4paper,fleqn]{cas-dc}

\usepackage[numbers]{natbib}
\usepackage{algorithm}
\usepackage{algpseudocode}
\usepackage{blindtext}
\usepackage{amssymb}
\usepackage{balance}
\usepackage{amsmath}
\usepackage{booktabs}
\usepackage{soul}
\usepackage[T1]{fontenc}
\usepackage{float}
\usepackage{hyperref}
\usepackage{enumitem}
\usepackage{tikzsymbols}
\usepackage{graphicx}
\usepackage{stfloats}

\def\tsc#1{\csdef{#1}{\textsc{\lowercase{#1}}\xspace}}
\tsc{WGM}
\tsc{QE}
\tsc{EP}
\tsc{PMS}
\tsc{BEC}
\tsc{DE}

\begin{document}

\begin{titlepage}
\begin{center}
\vspace*{10pt}
\textbf{\Large A Multimodal-Multitask Framework with Cross-modal Relation and Hierarchical Interactive Attention for Semantic Comprehension}
\vspace*{20pt}

Mohammad Zia Ur Rehman$^{a}$ (phd2101201005@iiti.ac.in)\\ Devraj Raghuvanshi$^b$ (devraj\_raghuvanshi@brown.edu)\\
Umang Jain$^a$ (cse200001076@iiti.ac.in)\\ Shubhi Bansal$^a$ (phd2001201007@iiti.ac.in)\\ Nagendra Kumar$^a$ (nagendra@iiti.ac.in) \\

\hspace{1pt}

\begin{flushleft}
$^a$ Indian Institute of Technology Indore, Madhya Pradesh India\\

$^b$ Brown University, United States\\

\vspace{2cm}
\normalsize
This is the preprint version of the accepted paper.\\
\textbf{Published in \textit{Information Fusion}, 2025} \\
Published version is available at: https://doi.org/10.1016/j.inffus.2025.103628 \\
or \\
Published version (Free Link): https://authors.elsevier.com/c/1lebG5a7-G-7gU

Free link is valid till: October 11, 2025.

\end{flushleft}        
\end{center}
\end{titlepage}

\let\WriteBookmarks\relax
\def\floatpagepagefraction{1}
\def\textpagefraction{.001}

\shorttitle{}

\shortauthors{Zia et~al.}

\title [mode = title]{A Multimodal-Multitask Framework with Cross-modal Relation and Hierarchical Interactive Attention for Semantic Comprehension}                      



%
\author[1]{Mohammad Zia Ur Rehman}[type=editor,
                        auid=000,bioid=1,
                        prefix=,
                        role=,
                        orcid=0000-0001-6374-8102]


\ead{phd2101201005@iiti.ac.in}


\credit{Conceptualization, Methodology, Software, Investigation,
Writing - Original Draft, Writing - review \& editing}

\affiliation[1]{organization={Indian Institute of Technology Indore},
    country={India}}

\author[2]{Devraj Raghuvanshi}
\ead{devraj_raghuvanshi@brown.edu}
\credit{Conceptualization, Methodology, Software, Investigation,
Writing - Original Draft, Writing - review \& editing}

\author[1]{Umang Jain}
\ead{cse200001076@iiti.ac.in}
\credit{Software, Formal analysis, Visualization, Writing - Original Draft}

\author[1]{Shubhi Bansal}
\ead{phd2001201007@iiti.ac.in}
\credit{Software, Formal analysis, Writing - Original Draft}

\author[1]{Nagendra Kumar}[type=editor,
                        auid=000,bioid=1,
                        prefix=,
                        role=,
                        orcid=]

\cormark[1]
\ead{nagendra@iiti.ac.in}

\credit{Conceptualization, Methodology, Supervision, Writing - review \& editing}

\affiliation[2]{organization={Brown University},
    country={United States}}



\cortext[cor1]{Corresponding author}



\begin{abstract}
A major challenge in multimodal learning is the presence of noise within individual modalities. This noise inherently affects the resulting multimodal representations, especially when these representations are obtained through explicit interactions between different modalities. Moreover, the multimodal fusion techniques while aiming to achieve a strong joint representation, can neglect valuable discriminative information within the individual modalities.
To this end, we propose a Multimodal-Multitask framework with crOss-modal Relation and hIErarchical iNteractive aTtention (MM-ORIENT) that is effective for multiple tasks. The proposed approach acquires multimodal representations cross-modally without explicit interaction between different modalities, reducing the noise effect at the latent stage. To achieve this, we propose cross-modal relation graphs that reconstruct monomodal features to acquire multimodal representations. The features are reconstructed based on the node neighborhood, where the neighborhood is decided by the features of a different modality. We also propose Hierarchical Interactive Monomadal Attention (HIMA) to focus on pertinent information within a modality. While cross-modal relation graphs help comprehend high-order relationships between two modalities, HIMA helps in multitasking by learning discriminative features of individual modalities before late-fusing them. Finally, extensive experimental evaluation on three datasets demonstrates that the proposed approach effectively comprehends multimodal content for multiple tasks. The code is available in the GitHub repository.\\
\href{}{https://github.com/devraj-raghuvanshi/MM-ORIENT}
\end{abstract}




\maketitle

\section{Introduction}

\label{sec:sa_int}
With the continuous evolution of online social media platforms, the practice of sharing multimodal content has become widespread \cite{9765342, MAJUMDER2018124}. 
Multimodal content refers to content that is presented using multiple forms of media, such as text, images, audio, and video. Combinations of different modalities, for instance, memes that contain text superimposed on images, enable the users to articulate different emotions from various perspectives, ultimately enhancing the overall communication experience. Memes are frequently used as a means to communicate humor, sarcasm, motivation, or even expressions of hate through a combination of text and visual elements. These visual and textual modalities provide distinct perspectives on a subject, thereby complementing each other and playing a vital role in the overall comprehension \cite{10068184,10123038}. Due to its ability to convey emotions and opinions in a concise and easily shareable format, multimodal content is widely shared on social media~\cite{rehman-etal-2025-implihatevid}.

The presence of an extensive collection of multimodal social media content has garnered attention toward the computational analysis of such content. Existing approaches have explored multimodal content within the field of sentiment analysis \cite{GANDHI2023424}, as well as the realms of hateful content identification \cite{3531925, CHHABRA2023106991}, sarcasm detection \cite{3124420} and humor recognition \cite{3551792}. Multimodal content can convey a range of emotions; hence, there is a pressing requirement to address multiple cognitions, giving rise to the advancement of multimodal-multitask learning.

\subsection{Existing Approaches and Challenges}
\begin{figure}[t!]
    \centering
    \includegraphics[width=8.7cm, height=13.5cm]{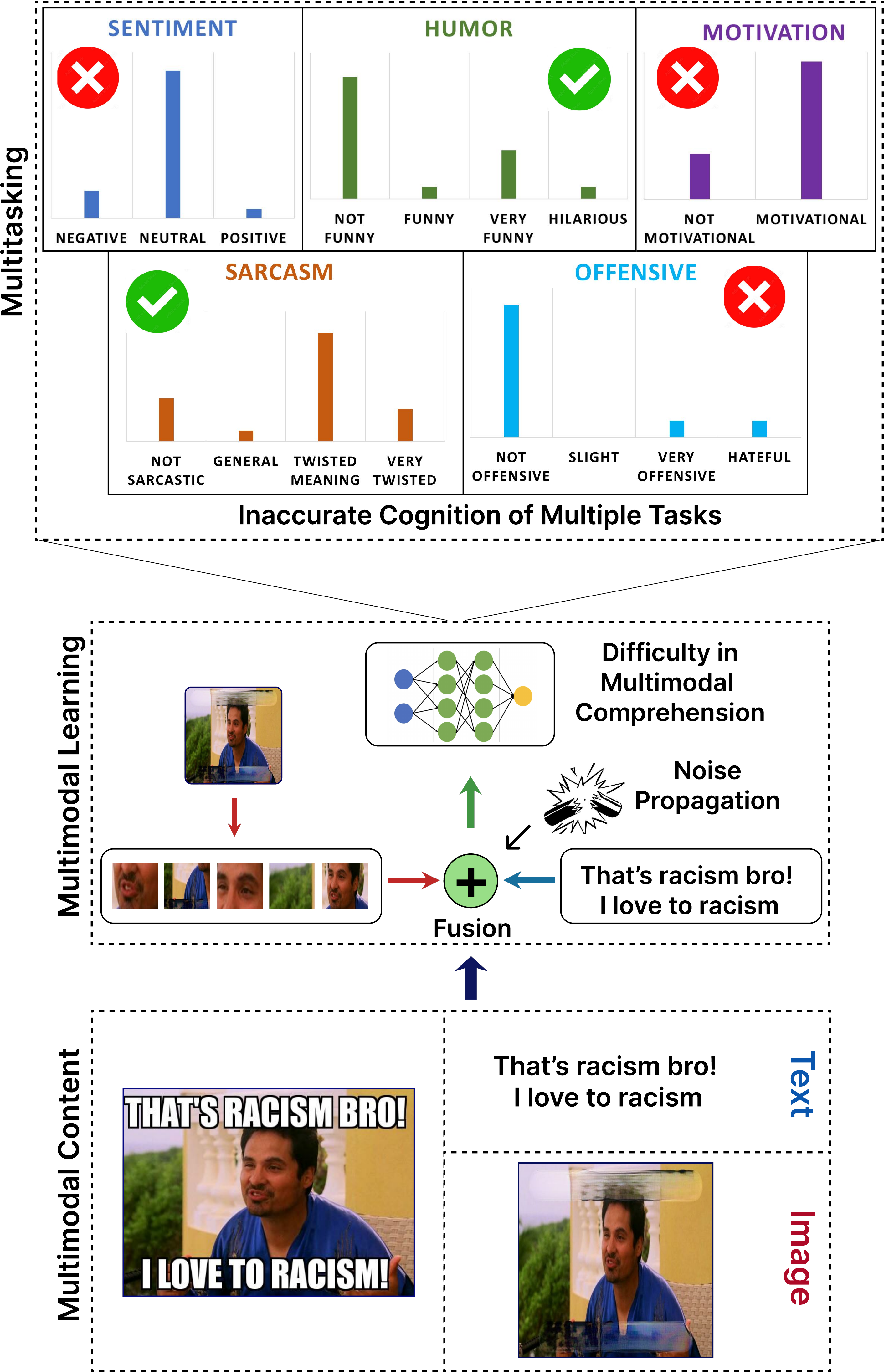}
    \caption{Multimodal Machine Cognition: Noise in monomodal features affects the comprehension of multimodal content, leading to inaccurate detection of different tasks.}
    \label{fig:intro}
\end{figure}

Prior research has often focused on creating fusion techniques for acquiring multimodal features from unimodal features \cite{9928434}. These fusion techniques encompass various approaches, including embedding fusion \cite{3618057}, transformer-based fusion \cite{tsai-etal-2019-multimodal}, and attention-based fusion \cite{9806458}. However, these techniques do not take into account the noise in monomodal features. Such noise, often arising from shared dominant components, anisotropy, or semantic redundancy, has been shown to obscure discriminative signals in embedding spaces \cite{mu2018all}. We define this embedding-level noise as the presence of irrelevant, low-variance, or misaligned components that hinder effective multimodal fusion. This is distinct from raw input noise, which typically originates from modality-specific distortions such as background clutter in images or typographical errors in text; while raw noise affects the input data directly, embedding-level noise emerges during representation learning and interferes with cross-modal alignment.
This inherent noise in monomodal representations influences cross-modal interactions and the overall multimodal representation \cite{9767641}, leading to inconsistent results. 
For instance, the meme displayed in the \autoref{fig:intro} is very offensive, has a negative sentiment, and is not motivational. However, it is misclassified as not offensive, neutral, and motivational. This is particularly critical in multimodal architectures involving cross-modal interaction mechanisms, such as cross-attention or co-attention, where representations from one modality are used to condition or modulate representations from the other modality through operations such as dot-product similarity. In such cases, noise in one modality's embedding space can propagate and directly contaminate the other modality’s features due to the multiplicative nature of these interactions. Such instances can be dealt with by cross-modal learning without explicit interaction between the two modalities. Hence, it is imperative to explore multimodal feature representation methods that do not require feature interaction explicitly, reducing the influence of noise while enhancing the representations in a cross-modal fashion. 

Earlier investigations have also established that one modality takes precedence in multimodal learning, with other modalities serving more as supplementary components \cite{mai-etal-2019-divide}. The existing fusion methods used to generate multimodal features lead to the oversight of discriminative information in monomodal features. 
Moreover, it is worth noting that memes have text superimposed on images, and this text sometimes lies on the important regions of the image, which hampers the extraction of pertinent features and creates noisy features. This necessitates removing text from the images in order to get refined and less noisy features. In the existing works, data augmentation has proven highly successful in enhancing model performance across various computer vision and natural language processing tasks \cite{ZIAURREHMAN2023103450}. However, the recent state-of-the-art generative AI-based large language models (LLMs) have not yet been explored for multimodal-multitask learning.

\subsection{Navigating Our Approach: Proposed Solution at a Glance }
In light of the aforementioned challenges, we propose a hierarchical interactive attention-aware multimodal-multitask framework for semantic comprehension with cross-modal relation learning. The proposed method learns latent multimodal features in a cross-modal manner without explicit interaction between different modality representations. This is achieved by constructing cross-modal relation graphs where nodes are created based on the features of one modality, and the edges between nodes are established based on the features of another modality. This method effectively reduces the impact of monomodal feature noise during cross-modal feature updates. The superimposed text region of a meme undergoes inpainting before feature extraction, effectively reducing the introduction of noise. We propose a \textbf{H}ierarchical \textbf{I}nteractive \textbf{M}onomodal \textbf{A}ttention (HIMA) mechanism. At the first level, HIMA generates region-based and word-level attention vectors; subsequently, at the next level, it creates a unified representation for the batch of each modality. These representations are late-fused to generate a joint representation. Cross-modal relation graphs help in multimodal comprehension, whereas the proposed attention mechanism benefits multitask learning. Additionally, 
we employ generative AI-based LLM for text augmentation.  
Lastly, task-specific text features are added in the final feature representation. The framework is primarily designed and evaluated on image-text multimodal content. However, its modular approach may be used to extend it for other modalities too.

\subsection{Key Contributions}
To sum up, our key contributions can be summarized as follows:
\begin{enumerate}
    \item{We propose a novel framework for semantic comprehension of multimodal content for multiple tasks, such as sarcasm detection, humor recognition, offensive content detection, motivation detection, and sentiment classification.} 
    \item{We design a novel feature reconstruction approach for semantic multimodal comprehension via cross-modal relation graphs. This approach does not necessitate a direct interaction between distinct modality representations, leading to a reduction in noise propagation in the latent multimodal features. }
    \item{We introduce a Hierarchical Interactive Monomodal Attention (HIMA) to benefit multitasking.  
    HIMA is applied on monomodal features in two steps: first, on the individual images and text, and later, on the batch of multimodal content. Unlike traditional Hierarchical Attention Network (HAN) \cite{yang-etal-2016-hierarchical}, which is applied on the document level and where the second stage vector (sentence vector) serves as the classification vector, HIMA utilizes both levels of vectors and finally integrates two modalities to generate a joint representation.} 
    \item{We evaluate the performance of the proposed method on three multimodal datasets. Quantitative and qualitative experiments conducted for five tasks provide confirmation of the efficacy of the proposed approach. } 
\end{enumerate}

\section{Related Work}

As Multimodal content combines textual and visual elements, there is a need for a comprehensive analysis of both components \cite{10123038} to identify underlying humor, sarcasm, or hostility effectively. 
Numerous approaches have been proposed in the existing works for multimodal emotion classification, encompassing graph-based methods, attention-based multimodal learning, and multitask learning. A few approaches employ data augmentation techniques to balance the data and increase the training samples.

\subsection{Graph-based Learning}
Lu et al. \cite{lu2024bi} introduce a Bi-stream Graph Learning-based Multimodal Fusion (BiGMF) framework specifically designed for Emotion Recognition in Conversation (ERC). Their method explicitly separates intra-modal contextual learning and inter-modal relational modeling using two parallel streams: unimodal stream graph learning with unimodal graph attention networks (UMGATs), and cross-modal stream graph learning with cross-modal graph attention networks (CMGATs). This design alleviates conflicts arising from heterogeneous modality fusion and ensures that contextual and complementary information are effectively preserved. Importantly, CMGATs enable direct inter-modality feature interaction at the graph level, and a novel cross-modal loss is introduced to further constrain and regularize inter-modal learning. This explicit modeling approach contrasts with prior works that relied primarily on feature concatenation, offering better interpretability and performance on benchmark datasets like IEMOCAP and MELD. Similarly, Li et al. \cite{li2024integrating} present a GIN-based multimodal feature transformation framework coupled with a voting strategy for irony-aware cyberbullying detection. The model employs Graph Isomorphism Networks (GINs) to transform joint BERT- and ViT-based text-image features into rich node embeddings, capturing latent structural relationships in multimodal data. Notably, the proposed framework achieved a significant boost in F1 and AUC metrics, highlighting the effectiveness of structural feature transformation via GNNs in real-world noisy multimodal contexts.

Lian et al. \cite{10008078} propose an approach designed for incomplete modalities for sentiment classification in conversational videos. 
The approach employs a pair of graph-based networks to capture dependencies related to the speaker and temporal aspects. Sun et al. \cite{10068184} devise a brain-inspired hypergraph method to bridge the gap between multimodal and multitask learning to detect various emotions in memes. The method integrates monomodal, multimodal, and multitask hypergraph networks, each serving to replicate the brain processes of decomposition, association, and synthesis. 
Li et al. \cite{10078161} propose a graph-based multimodal emotion recognition approach. The approach is based on a pair-wise cross-modal complementation strategy to extract diverse edge types, enhancing the ability of Graph Neural Networks (GNNs) to capture essential contextual and interactive information. A few works \cite{10149528, 9712249,10008205} present efficient graph-based approaches for other downstream tasks such as movie genre prediction, micro-video recommendation, Amazon co-purchase graphs, and citation networks.

In the aforementioned multimodal methods, the graph-based methods utilize monomodal features from multiple modalities to create graphs. Monomodal representations include noisy data that adversely impacts the learning of cross-modal interactions in graphs \cite{9767641}. Furthermore, these techniques are not well-suited for complex combined tasks of multimodal-multitask learning. In contrast, we construct cross-modal relation graphs using features from one modality for nodes, while an edge between two nodes is created based on features of another modality. This approach mitigates the influence of monomodal feature noise while updating the node features in a cross-modal manner.

\subsection{Attention-based Multimodal and Multitask Learning}

Hossain et al. \cite{hossain2024align} propose a context-aware alignment framework that aligns visual and textual features before fusion, rather than relying on standard cross-attention. Their MCA-SCF model computes modality-guided context vectors using attention, which are then fused along with residual modality features. This strategy improves hateful meme detection performance on both English (MultiOFF) and Bengali (MUTE) datasets and demonstrates robustness in low-resource settings.
Yadav and Singh \cite{yadav2024hatefusion} present HateFusion, an explainability-enhanced system to detect implicit hate speech using fine-tuned BERT and DistilBERT. Their model employs attention mechanisms for classification and integrates interpretability tools such as LIME and SHAP to highlight influential tokens. HateFusion shows strong performance across binary, 3-way, and 7-way tasks, and includes a GUI for real-time feedback and flagging, enabling dynamic model refinement.

Yao et al.\cite{9482580} propose a multimodal, multiinteractive, and multihierarchical neural network utilizing insights from neuroanatomy and neuropsychology for sarcasm detection in images and text. The method incorporates gating and guided attention for feature interaction. Zhang et al. \cite{ZHANG2023282} propose a multimodal-multitask framework for sarcasm detection utilizing single and multi-level decoders. This approach employs intramodal and intermodal attention to comprehend contextual dependencies among text, visual, and acoustic features. Ghosh et al. \cite{ghosh-etal-2022-comma} design a commonsense-aware framework for emotion recognition for multimodal mental health conversations. This method integrates commonsense reasoning and devises a gating mechanism to eliminate unwanted noise. Lv et al. \cite{Lv_2021_CVPR} observe that the lack of synchronization between modalities adds complexity to the task of achieving effective multimodal fusion. Consequently, the study devises an approach to handle the issue of asynchrony of modalities in multimodal streams for sentiment and emotion detection. 
A few works use a transformer as the backbone for multimodal hate content detection \cite{Bhandari_2023_CVPR} and multimodal representation for languages \cite{10005816}. 

The aforementioned works mostly rely on the attention mechanism for feature interaction among different modalities. However, feature interaction through a cross-attention mechanism may result in inter-modal incongruity due to different learning dynamics between modalities and diverse noise patterns.

\subsection{Data Augmentation Methods}
Data augmentation techniques involve enhancing the training data by introducing variations and transformations across multiple data modalities. Various techniques have been used for text classification, such as the Easy Data Augmentation (EDA) of random swap and deletion method \cite{wei-zou-2019-eda}, word replacement \cite{li-etal-2017-robust}, and replacement by language models like BERT \cite{jiao-etal-2020-tinybert}.
For images, transformations like rotations and adjustments to the RGB channel can be valuable, as the desired outcome is a model that remains invariant to these changes. Image augmentation has proven highly effective in a wide range of domains, including emotion and sentiment classification \cite{10250883}, image classification \cite{He_2016_CVPR}, and object detection \cite{Zhong_Zheng_Kang_Li_Yang_2020}. We utilize the state-of-the-art generative AI method GPT-3.5 for text augmentation and the randomized transformation process for image augmentation.

\section{Problem Definition and Objectives}

The primary challenges arise from the complex interplay between text and images. 
Furthermore, the multitask nature of the problem introduces additional complexity, as each meme may convey multiple emotions simultaneously.

\subsection{Main Problem}
Given a dataset with $N$ samples, \(D = \{(X_i, Y_i)\}_{i=1}^N\), where $(X_i, Y_i)$ represents a multimodal sample with aligned text and image modalities $X = (x_i^{\text{txt}}, x_i^{\text{img}})$ and associated labels  $Y_i \in \{y_1, y_2,..., y_n \}$ where each $y_i$ denotes a single task. Our goal is to build a unified framework \(F\) that effectively comprehends the multimodal content $X_i$ and outputs the multitask labels $Y_i$ as follows:

\begin{align}
    F(X_i) &= Y_i \label{eq:prob1} \\
    \text{s.t. } F(x_i^{\text{txt}}, x_i^{\text{img}}) &= (y_1, y_2, ...., y_n)\label{eq:prob2}
\end{align}

We define two major subproblems as our objectives, each addressing a key aspect of our framework.
\subsection{Objective 1: Cross-modal Feature Interaction}

Given \(E_i^{m_1}\) and \(E_i^{m_2}\), denoting the features extracted from modalities \(m_1\) and \(m_2\) respectively, where $m_1, m_2 \in \{\text{txt}, \text{img}\}$, our objective is to generate refined feature representations for a given task using cross-modal relation graphs \(\mathcal{G}_{\text{cross}}(m_1,m_2)\) wherein $m_1 \neq m_2$, nodes correspond to features of the \(m_1\) modality, while edges are derived from the features of the \(m_2\) modality. This approach is designed to mitigate the impact of monomodal feature noise, thereby enhancing the representation of multimodal data. Mathematically, this can be expressed as:

\begin{equation}
\mathcal{R}_i^{m_1} = \begin{cases}
        \begin{aligned}
                & f(E_i^{\text{txt}}, \mathcal{G}_{\text{cross}}(\text{txt},\text{img})) , m_1=\text{\text{txt}}\\
                & f(E_i^{\text{img}}, \mathcal{G}_{\text{cross}}(\text{img},\text{txt})), m_1 = \text{\text{img}}
        \end{aligned}
    \end{cases}
\end{equation}

where \(\mathcal{R}_i^{m_1}\) represents the refined feature representation of the \(i\)-th sample in \(m_1\) modality, \(f\) is the function that generates refined feature representations for $E_i^{m_1}$ based on \(\mathcal{G}_{\text{cross}}(m_1,m_2)\).

\subsection{Objective 2: Hierarchical Attention-based Feature Interaction}

 For a given modality $m \in \{\text{txt}, \text{img}\}$, our objective is to develop a function \(\mathcal{H}^m(Q_i^m, \gamma^m)\) that effectively harnesses the monomodal features at different levels of granularity to produce an enhanced joint feature representation for a given sample. Here, $m \in \{\text{txt}, \text{img}\}$ represents the modality. Mathematically, this can be formulated as:

 \begin{equation}
 \mathcal{H}^m = \begin{cases}
    \begin{aligned}
          & \mathcal{H}^{\text{txt}}(f_{\text{word}}(Q_i^{\text{txt}}), f_{\text{sentence}}(\gamma^{\text{txt}})), m=\text{txt} \\
          & \mathcal{H}^{\text{img}}(f_{\text{region}}(Q_i^{\text{img}}), f_{\text{image}}(\gamma^{\text{img}})), m=\text{img}
    \end{aligned}

 \end{cases}
 \end{equation}

where \(Q_i^m \in \gamma^m\) represents the feature matrix of $m$ modality associated with the \(i\)-th sample, and \(\gamma^m\) is a batch of feature matrices corresponding to $m$ modality. The functions \(\mathcal{H}^{\text{txt}}\) and \(\mathcal{H}^{\text{img}}\) process the outputs of the attention mechanisms. \(f_{\text{word}}(Q_i^{\text{txt}})\), \(f_{\text{sentence}}(\gamma^{\text{txt}})\), \(f_{\text{region}}(Q_i^{\text{img}})\), and \(f_{\text{image}}(\gamma^{\text{img}})\) are attention mechanisms that operate at different levels of granularity within each modality, with their corresponding levels mentioned in the subscripts.

\section{Methodology}
\begin{figure*}[H]
    \centering
    \includegraphics[width=1\linewidth]{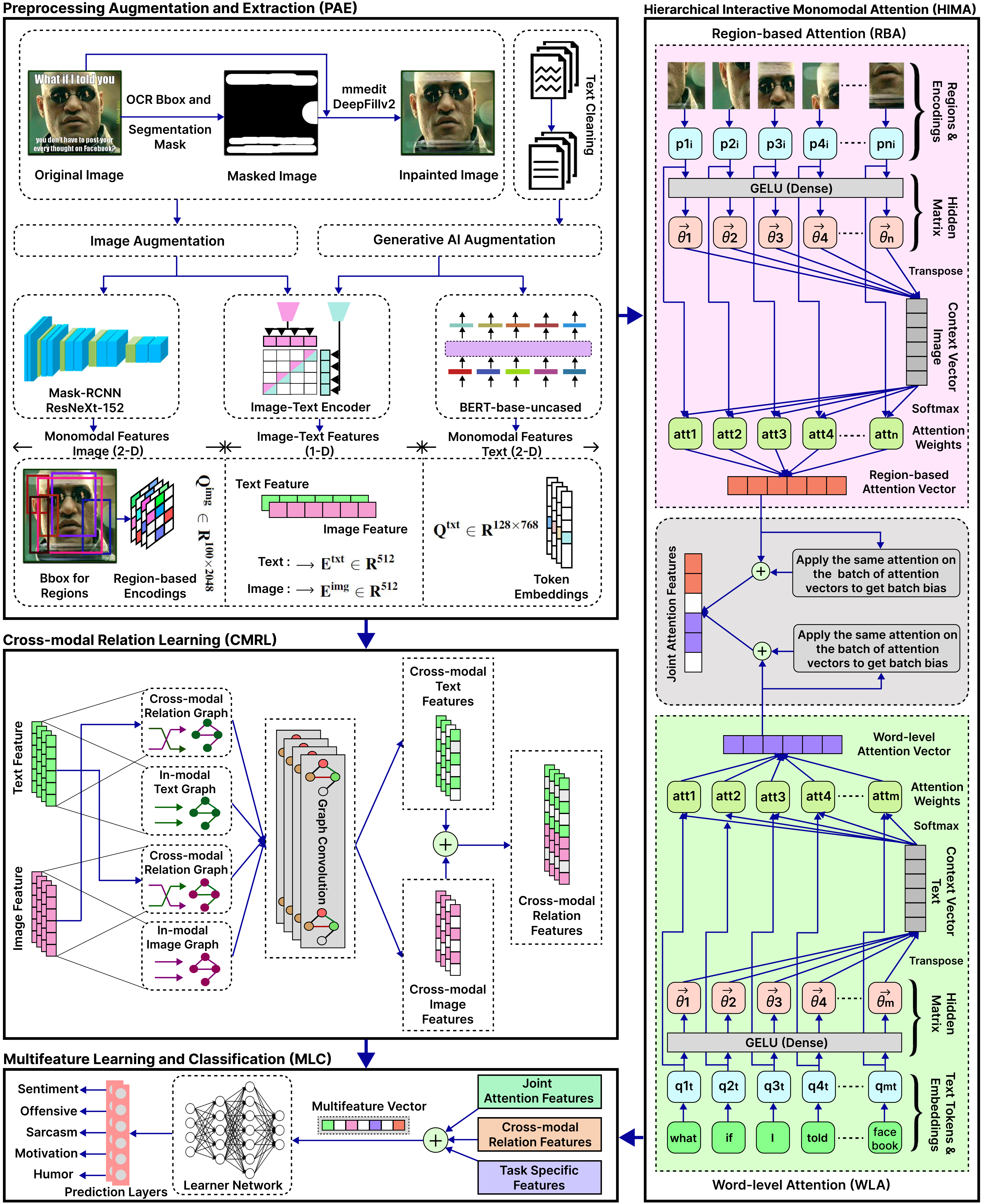}
    \caption{The figure represents the overall architecture of the proposed system. Input to the PAE module is a meme that consists of image and text modalities. PAE module provides textual and visual features as output. The HIMA module takes monomodal features as input and outputs an attended joint representation. The CMRL module takes as input the features extracted from a multimodal encoder and provides cross-modal relation features as output. The MLC module consists of the learner network, which takes inputs from CMRL and HIMA modules. It is additionally provided with text-based emotion, toxicity, and sentiment features. \autoref{fig:in-modal} and \autoref{fig:cross-modal} present In-modal and Cross-modal relation graph construction in detail.}
    \label{fig:system_architechture}
\end{figure*}

In this section, we describe the methodology of our proposed multimodal-multitask framework, MM-ORIENT, for semantic comprehension. As depicted by the \autoref{fig:system_architechture}, the proposed framework first preprocesses the images and text and then extracts the features for the HIMA and Cross-modal Relation Learning (CMRL) modules. Finally, a joint representation of the features is passed through the learner network for emotion detection. Subsequent subsections elaborate on the proposed approach.

\subsection{Data Preprocessing}
Data preprocessing entails two components: Image Masking \& Inpainting and Text Cleaning. Image masking \& inpainting ensures the removal of overlaid text to emphasize visual elements, while text cleaning standardizes text format for accurate extraction.

\subsubsection{Image Masking \& Inpainting}
Memes often contain humorous or contextual text overlaid on the image. This text can introduce noise and interfere with the feature extraction process. To eliminate the text from image, we utilize masking and inpainting techniques as shown in \autoref{fig:system_architechture}. This ensures that the primary focus is on the underlying visual content of the image, allowing for more precise and pertinent feature representations.

Masking involves the creation of a binary image mask, separating text from the underlying visual content, and enabling the removal of text-containing regions in the image. This allows for subsequent inpainting of the text-removed regions that involves generating substitute elements in areas that are absent, ensuring that the alteration appears both visually convincing and semantically accurate. We employ a generative image inpainting technique based on gated convolutions proposed by Yu et al. \cite{yu2019free} to produce the absent sections within masked images, thereby achieving a natural appearance. By using masking in conjunction with inpainting techniques, we are able to remove the text from the image entirely.

\subsubsection{Text Cleaning}
Text cleaning encompasses tasks like purging undesirable characters, symbols, and excessive white spaces, as well as standardizing its case. If URLs or usernames are present in the image, they are removed to ensure only relevant textual content is retained. This step is essential to ensure that the extracted text is in a clean and standardized format, which in turn improves the quality of extracted features.

While these preprocessing steps help improve input quality, the core contribution of our work lies in mitigating structural noise in the learned feature space, which impacts multimodal representation learning.

\subsection{Data Augmentation}

Data augmentation enriches our dataset with diverse variations and increases the minority class samples, thus enhancing the adaptability of our models for finer interpretation of multimodal content. For each image and text in the training data, we generate a few more similar images and text. These new images and text are added to the training data before the training of the model.

\subsubsection{Image Augmentation}
For image augmentation, we strategically manipulate image attributes by subjecting each input image to a randomized transformation process. This involves horizontal flipping, which mirrors the image along the vertical axis, and adjustments to brightness, contrast, hue, and saturation. The range of these adjustments is carefully selected to introduce meaningful variations while preserving the underlying content. Hence, each image is transformed using one of these transformations that are selected randomly. The resultant image is added to the training set and is given the same label as the original. The text of the given image is also transformed using text augmentation.

\subsubsection{Text Augmentation}

For text augmentation, we harness the capabilities of OpenAI's state-of-the-art language model, GPT-3.5-turbo, using a technique known as few-shot prompting. This method allows us to provide specific examples that guide the model's responses in generating alternative versions of sentences. By framing the task as a conversational exchange, we prompt the system to produce several distinct rephrased renditions for each original sentence. We prepare the following prompt for the GPT-3.5; \textit{\{``role": "system", "content": "You are a helpful, pattern-following assistant that removes the URLs and usernames from sentence, and then rephrases the sentence in 5 different forms"\}. Subsequently, sentences are provided to the GPT-3.5 which returns sentences in different words without losing the meaning of the text.} This approach enables us to diversify our dataset with a curated set of linguistically diverse alternatives, ultimately enhancing the richness and variability of our training corpus.

\subsection{Feature Extraction}
Within our framework, we incorporate various feature extraction techniques for image and text modalities. 
Additionally, we introduce supplementary features like emotion, sentiment, and toxicity-related attributes derived from the text. 

\subsubsection{Feature Extraction for CMRL}
We utilize the pre-trained CLIP \cite{radford2021learning} encoder for extracting features from datasets having both image and text modalities in a unified manner. Unlike traditional methods that focus on either images or text separately, it is trained to learn representations that can bridge the gap between the two modalities. It learns to associate images with their corresponding text, allowing it to generate meaningful representations for both modalities altogether. The encoder processes the image-text pair $X_i$ and produces image embeddings $E^{\text{img}} \in \mathbb{R}^{512}$ and text embeddings $E^{\text{txt}} \in \mathbb{R}^{512}$. 
\begin{equation}
    E^{\text{img}}, E^{\text{txt}} = \overrightarrow{\text{CLIP}}(X_i)
\end{equation}

Compared to other multimodal models, CLIP produces embeddings with lower dimensionality. The efficiency of CLIP in generating embeddings quickly and at scale aligns well with the requirements of GraphSAGE, which benefits from rapid and scalable embedding computations making it an ideal choice.

\subsubsection{Feature Extraction for HIMA}
We employ BERT \cite{devlin2018bert} for textual feature extraction. The \texttt{bert-base-uncased} model is used for tokenization, breaking down raw text into tokens. For each sample in the training data, the tokenized input is passed through BERT. The model is configured with a maximum sequence length of 128 tokens, enabling truncation and padding for uniform sizes. The BERT model processes the set of tokens $\{w_i\}_{i=1}^{128}$, producing the feature matrix $Q^{\text{txt}}=\{q_i\}_{i=1}^{128}$, where $q_i \in \mathbb{R}^d$ represents the token embedding of the $i$-th token and d denotes the dimention of token embedding. For $m=$ txt, $d=768$.

\begin{equation}
    q_i = \overrightarrow{\text{BERT}}(w_i)
\end{equation}

For the visual modality, we employ a region-based image feature extraction approach detectron2~\cite{wu2019detectron2} utilizing a Mask RCNN \cite{he2017mask} model with ResNeXt-152 as the backbone network. We refer to this model as Mask RCNN X152. Each image is divided into 100 regions $\{r_i\}_{i=1}^{100}$, from each of which a feature vector $p_i \in \mathbb{R}^d$ is extracted where $d=2048$. These features are derived from the \texttt{fc6} layer of the ResNeXt-152 backbone network. The resulting visual feature matrix is denoted as $Q^{\text{img}} = \{p_i\}_{i=1}^{100}$.

BERT has achieved state-of-the-art results on a wide range of NLP benchmarks, demonstrating its effectiveness in capturing linguistic insights and improving the performance of downstream tasks such as text classification. Detectron2 generates precise region proposals, identifying multiple objects within an image and delineating their boundaries accurately. This is particularly useful for extracting detailed region-specific features. Detectron2 extracts rich, multi-scale features from images, capturing fine details and high-level semantics.

\begin{equation}
    p_i = \overrightarrow{\text{MRCNN-X152}}(r_i)
\end{equation}

\subsubsection{Task Specific Features}
\label{sec:task-specific-features}
We incorporate emotion features, sentiment features, and toxicity-related features to enhance the text analysis capabilities of MM-ORIENT. These features provide a comprehensive understanding of the underlying sentiment and emotional insights within the text samples.
To extract emotion features, each word in the text is mapped to its underlying emotion category \{fear, anger, anticipation, trust, surprise, positive, negative, sadness, disgust, joy\}. We then obtain a feature vector for the given text sample, where each element corresponds to the frequency of a specific emotion category\footnote{\label{nrclex}\href{https://pypi.org/project/NRCLex/}{https://pypi.org/project/NRCLex/}}.
 
Furthermore, we retrieve sentiment values for each text sample. The sentiment categories span from 0 to 4, encompassing Very Negative, Negative, Neutral, Positive, and Very Positive sentiment categories~\cite{manning2014stanford}.
To extract toxicity-related features, we leverage the RoBERTa Toxicity Classifier\footnote{\label{roberta_toxicity}\href{https://huggingface.co/s-nlp/roberta_toxicity_classifier}{https://huggingface.co/s-nlp/roberta\_toxicity\_classifier}}, which is trained to classify the text in toxic and not-toxic classes.
A 768-dimensional feature vector is extracted for toxicity elements.

\subsection{Hierarchical Interactive Monomodal Attention}

Hierarchical attention-based feature interaction has proved to be very effective in monomodal as well as multimodal tasks \cite{9765342,yang-etal-2016-hierarchical}. Understanding the intricate and subtle interplay of words within text modality and regions in image modality is paramount for accurately discerning the contextual meaning of a given meme. Hence, inspired by Rehman et al.~\cite{REHMAN2025126285}, we employ word-level attention (WLA) and region-based attention (RBA) mechanisms designed to emphasize pivotal words and regions within the text and image modality, respectively, enabling the model to synthesize a comprehensive understanding of memes, capturing both linguistic subtleties and visual cues for a more accurate interpretation of their intended message. Algorithm \ref{alg-attention} shows the implementation of HIMA.

\begin{algorithm}
    \caption{Hierarchical Interactive Monomodal Attention}
    \label{alg-attention}
    \begin{tabular}{ll}
    \textit{Input:} & $k$: Index of a sample in the batch\\
    & $\gamma^m$: Feature batch containing $k$-th sample\\
  \textit{Output:} 
  & $z_k^m$: Multi-level contextual vector for $k$-th sample\\
    \end{tabular}
\begin{algorithmic}[1]
\Function{Hierarchical\_Attention}{$k, \gamma^m$}
            \ForAll{$Q_y^m \in \gamma^m$}
                    \ForAll{$v_x^m \in Q_y^m$}
                        \State ${\theta}_x^m \gets \text{GELU}(v_x^m W_1^m + b_1^m)$
                    \EndFor
                    \State ${hid\_rep1}_y^m \gets \{{\theta}_x^m\}_{x=1}^L$
                    \ForAll{${\theta}_x^m \in {hid\_rep1}_y^m$}
                        \State ${att}_x^m \gets \text{softmax}(({\theta}_x^m)^T u_1)$
                    \EndFor
                    \State $t_y^m \gets \sum_{x=1}^{L} {att}_x^m v_x^m$
            \EndFor
            \State $T^m \gets \{t_y^m\}_{y=1}^B$
                    \ForAll{$t_y^m \in T^m$}
                        \State $l_y^m \gets \text{GELU}(t_y^m W_2^m + b_2^m)$
                    \EndFor
                    \State ${hid\_rep2}_y^m \gets \{l_y^m\}_{y=1}^B$
                    \ForAll{$l_y^m \in {hid\_rep2}_y^m$}
                        \State $p_y^m \gets \text{softmax}((l_y^m)^T u_2)$
                    \EndFor

                    \State $s^m \gets \sum_{y=1}^{B} p_y^m t_y^m$
                    \State $z_k^m \gets \text{concat}(t_k^m, s^m)$
                    \State \textbf{return} $z_k^m$
        \EndFunction
    \end{algorithmic}
\end{algorithm}

Suppose we have  $\gamma^m = \{Q_y^m\}_{y=1}^B$ which denotes a batch of 2-dimensional features extracted from $m \in \{\text{txt}, \text{img}\}$ modality, where $Q_y^m = \{v_x^m\}_{x=1}^{L}$ represents the $y$-th feature matrix of $m$ modality that corresponds to a single sample in the batch, and $B=128$ denotes the batch size. Here, $v_x^m \in \mathbb{R}^d$ represents the word/region embedding vector. For $m=\text{txt}$, $v_x^m$ represents the word embedding of the $x$-th word in the sequence, and $L$ denotes the sequence length. For $m=\text{img}$, $v_x^m$ represents the embedding vector of the $x$-th bounding box or region in the image, and $L$ denotes the total number of regions.

\subsubsection{Word-level and Region-based Attention (stage 1)}
Both WLA and RBA follow the same procedure. First, we compute the hidden representation \({\theta}_x^m\) of $v_x^m$ through the transformation using the Gaussian Error Linear Unit (GELU), which provides a smooth approximation to the Rectified Linear Unit (ReLU) while maintaining smooth gradients:

\begin{equation}
    \label{eq-attention:1}
    {\theta}_x^m = \frac{1}{2} \left(v_x^m W_1^m + b_1^m\right) \left(1 + \text{erf}\left(\frac{v_x^m W_1^m + b_1^m}{\sqrt{2}}\right)\right)
\end{equation}

where \texttt{erf} is the error function, the $W$ is the weight parameter and $b$ is the bias. $W_1^m \in \mathbb{R}^{768 \times \beta}$ and $b_1^m \in \mathbb{R}^\beta$ are fine-tuned during training. Here, $\beta$ denotes the output dimension of the linear transformation. We evaluate the significance of the $x$-th word/region by calculating the similarity between ${\theta}_x^m$ and a word/region-level context vector $u_1$. Subsequently, we obtain the attention weights ${att}_x^m$ using a softmax function:

\begin{equation}
    \label{eq-attention:2}
    {att}_x^m = \frac{\exp(({{\theta}_x^m})^T u_1)}{\sum_{x=1}^{L} \exp(({{\theta}_x^m})^T u_1)}
\end{equation}

These attention weights serve as dynamic coefficients, emphasizing words for $m=\text{txt}$ and regions for $m=\text{img}$ that carry heightened contextual significance, enabling the model to allocate greater importance to semantically relevant elements during the aggregation process. Finally, the WLA vector \(t_y^{\text{txt}}\) is computed, aggregating the contributions of individual words, and the RBA vector $t_y^{\text{img}}$ is computed, aggregating the contributions of individual regions, based on their respective attention scores:

\begin{equation}
    \label{eq-attention:3}
    t_y^m = \sum_{x=1}^{L} {att}_x^m v_x^m    
\end{equation}

\subsubsection{Sentence-level and Image-level Attention (stage 2)} 
Similarly, as we calculated \(t_y^m\) to capture significant cues at the word/region level, we extend our attention mechanism to a higher level to capture batch-wise contextual subtleties at the sentence/image level by introducing a sentence/image-level context vector $u_2$. Given a batch of WLA/RBA vectors $T^m=\{t_y^m\}_{y=1}^B$, where $t_y^m$ represents the WLA/RBA vector of the $y$-th sample in the batch, we calculate the attention vector $s^m$ in a similar manner as shown in Equations \ref{eq-attention:1}-\ref{eq-attention:3} with the only difference being that we use $t_y^m \in T^m$ instead of $v_x^m \in Q_y^m$.

\begin{equation}
l_y^m = \frac{1}{2} \left(t_y^m W_2^m + b_2^m\right) \left(1 + \text{erf}\left(\frac{t_y^m W_2^m + b_2^m}{\sqrt{2}}\right)\right)
\end{equation}

\begin{equation}
    p_y^m = \frac{\exp(({l_y^m})^T u_2)}{\sum_{y=1}^{n} \exp(({l_y^m})^T u_2)}
\end{equation}

\begin{equation}
    s^m = \sum_{y=1}^{B} p_y^m t_y^m    
\end{equation}

We refer to this attention vector $s^m$ as the batch bias that summarizes the information of all the sentences in the batch (for $m=\text{txt}$) or all the images in the batch (for $m=\text{img}$). In order to obtain a combined feature representation $z_k^m$ for the $k$-th sample in the batch, that encapsulates a multi-level or hierarchical contextual understanding of $m$ modality, we concatenate $t_k^m$ and $s^m$:

\begin{equation}
    z_k^m = \text{{concat}}(t_k^m, s^m)
\end{equation}

Finally, a joint attention feature vector $Z_k$ for the $k$-th meme is obtained by late-fusing $z_k^m$ obtained from both modalities:

\begin{equation}
    Z_k = \text{concat}(z_k^\text{txt}, z_k^\text{img})
\end{equation}

The $for$ loops in lines 2-3 of Algorithm \ref{alg-attention} will run for $B \times L$ iterations in total. In line 4, there is a matrix multiplication operation between two matrices with shapes $1 \times d$ and $d \times \beta$ contributing $O(d \times \beta)$. The remaining steps do not significantly impact the overall time complexity. Thus, the overall time complexity of Algorithm \ref{alg-attention} is $O(B \times L \times d \times \beta)$.

\subsection{Cross-modal Relation Learning}

The idea behind CMRL is to learn latent multimodal features through a cross-modal approach, without the need for direct interaction between representations of different modalities at the early stage. In order to accomplish this, we propose cross-modal relation graphs, which reconstruct monomodal features to obtain multimodal representations. The reconstruction of features relies on the node neighborhood, with the neighborhood determined by the features of a distinct modality. Algorithm \ref{alg-graph} shows the implementation of CMRL. The subsequent sections elaborate on the graph construction and feature reconstruction process.

\begin{algorithm}
    \caption{Cross-modal Relation Learning}
    \label{alg-graph}
    \begin{tabular}{ll}
    \textit{Input:} & $k$  : Index of a sample in the batch\\
    &  $X^{\text{txt}}$  : Text embedding matrix \\
    &  $X^{\text{img}}$  : Image embedding matrix \\
    & $thr$ : Similarity threshold \\
  \textit{Output:} 
  & $H_k$: Cross-modal relation feature vector\\

    \end{tabular}
    \begin{algorithmic}[1]
        \Function{Graph\_Learning}{$k, X^{\text{txt}}, X^{\text{img}}, threshold$}
           
            \ForAll{$m \in \{\text{txt}, \text{img}\}$}
                
                \State ${{norm\_X}}^m = {X^m}/{\| {X^m} \|_2}$
                \State $S^m \gets \{\text{cs}({\mathcal{E}}_i^m, {\mathcal{E}}_j^m)\} \forall \mathcal{E}_i^m, \mathcal{E}_j^m \in {{norm\_X}}^m\}$
                \State $A^m \gets \{A_{ij}^m | A_{ij}^m = 1 \textbf{ if } S_{ij}^m \geq threshold \textbf{ else } 0\}$
            \EndFor
            \State $G \leftarrow \{\}$
            \ForAll{$m_1 \in \{\text{txt}, \text{img}\}$}
                \ForAll{$m_2 \in \{\text{txt}, \text{img}\}$}
                    \State $G.\text{add}(\mathcal{G}(m_1,m_2))$
                \EndFor
            \EndFor
            \ForAll{$\mathcal{G} \in G$}
                \State $N \leftarrow \{\text{$V_i$: $V_i$ is a neighbour of the $k$-th node in $\mathcal{G}$}\}$
                \State $sum \leftarrow \sum_{V_i \in N} {E_i^{m_1}}$
                \State $M \gets sum/|N|$
                \State ${H}_k^{m_1\text{-}m_2} \gets \text{concat}(M, E_k^{m_1}) \cdot W + b$
            \EndFor
        \State ${H_k} \gets \text{concat}(H_k^{\text{img-img}}, H_k^{\text{txt-txt}}, H_k^{\text{img-txt}}, H_k^{\text{txt-img}})$
        \State \textbf{return} ${H_k}$
    \EndFunction
    \end{algorithmic}
\end{algorithm}

\subsubsection{Graph Construction}
The construction of the graph involves cosine similarity computation and adjacency matrix formation.

Given an embedding matrix \({X^m}=\{E_i^m\}_{i=1}^\lambda\) such that \(X^m \in \mathbb{R}^{\lambda \times D}\), where \(m \in \{\text{txt}, \text{img}\}\) is the modality, \(\lambda\) is the total number of embeddings, and $D=512$ is the embedding dimension, the embeddings are first normalized using L2 normalization to obtain a matrix ${{norm\_X}}^m = \{\mathcal{E}_i^m\}_{i=1}^\lambda$:

\begin{equation}
    {{norm\_X}}^m = \frac{{{X^m}}}{{\| {{X^m}} \|_2}}
\end{equation}

Here, $\|{X^m}\|_2$ represents the L2 norm of the embeddings, computed along axis 1, ensuring that each embedding vector has a unit length, ${{norm\_X}}^m$ denotes the normalized embeddings matrix, and $\mathcal{E}_i^m$ represents a single normalized embedding.

Next, the similarity matrix $S^m$ is computed using cosine similarity:

\begin{equation}
    S^m = \frac{{{norm\_X}^m \cdot {norm\_X}^m}}{{\| {norm\_X}^m \|_2 \cdot \| {norm\_X}^m \|_2}}
\end{equation}

Here, $S^m$ represents the cosine similarity between normalized embeddings. In the algorithm, ``cs" denotes the cosine similarity function. The resulting similarity matrix $S^m$ is a square matrix of size $\lambda \times \lambda$. Finally, a binary adjacency matrix $A^m$ is derived by thresholding $S^m$ at a specified $threshold$, indicating the presence or absence of edges between nodes:

\begin{equation}
    {A}_{ij}^m =
    \begin{cases}
    1, & \text{if } {S}_{ij}^m > {threshold} \\
    0, & \text{otherwise}
    \end{cases}
\end{equation}

The value of $A_{ij}^m$ indicates whether there is an edge between the $i$-th and $j$-th nodes, with 1 indicating the presence of an edge and 0 denoting its absence. 

We employ in-modal~\cite{REHMAN2025103895} and cross-modal relations by constructing a set of four graphs $G = \{\mathcal{G}(m_1,m_2)\}$ using all possible combinations of $m_1$ and $m_2$, where $m_1,m_2 \in \{\text{txt}, \text{img}\}$. $\mathcal{G}(m_1,m_2)$ represents a graph in which the edges are determined using the adjacency matrix $A^{m_2}$ and the nodes represent the embedding vectors of $m_1$ modality. Thus, we get two graphs representing in-modal relations ($m_1 = m_2$), and two graphs representing cross-modal relations ($m_1 \neq m_2$) as shown in \autoref{fig:in-modal} and \autoref{fig:cross-modal}, respectively. In the in-modal graphs, edges are computed based on the same modality as the nodes. For instance, if we are constructing an in-modal graph of text, then nodes are the text features, and edges are based on the similarity of the text features. Based on our experiments, we set the threshold values as 0.7 and 0.8 for constructing $\mathcal{G}(\text{txt},\text{txt})$ and $\mathcal{G}(\text{img},\text{img})$, respectively. Similarly, we set a threshold of 0.85 for $\mathcal{G}(\text{txt},\text{img})$ and 0.75 for $\mathcal{G}(\text{img},\text{txt})$.

\subsubsection{Feature Reconstruction}

Reconstruction of visual and textual features involves updating nodes by leveraging local neighborhood information via inductive representation learning. Inductive learning methods offer the benefit of generating embeddings for nodes or subgraphs that have not been previously seen. To employ inductive learning, we utilize GraphSAGE~\cite{hamilton2017inductive} that operates on node features and utilizes a sparse adjacency matrix to aggregate information from a node's neighborhood, enhancing the model's ability to capture complex relationships within the graph.
\begin{figure}[h]
    \centering
    \includegraphics[width=7 cm]{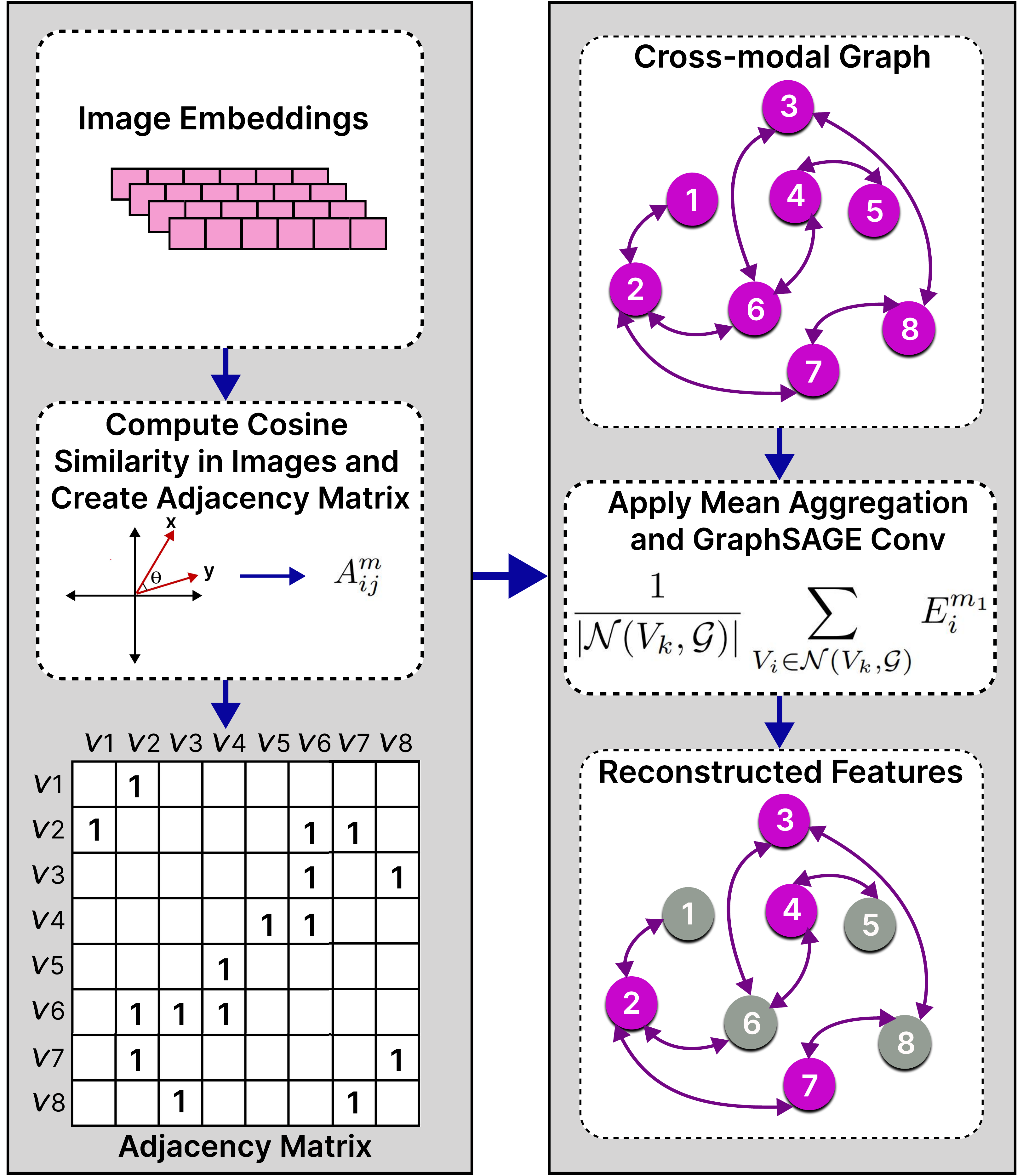}
    \caption{In-modal relation graph construction and updating: This graph represents the In-modal relation graph of an image. The vertices are image features, and an edge between two vertices is constructed based on cosine similarities of image features. A similar approach is used to create the In-modal text graph.}
    \label{fig:in-modal}
\end{figure} 
In the model architecture, the adjacency matrices ${A^\text{img}}$ and ${A^\text{txt}}$ are utilized as sparse inputs. The aggregation process is performed using the mean function. This function calculates the average of the neighboring nodes' features for each node, effectively creating a summarization of the local neighborhood information. The mean value for the $k$-th node $V_k$ of a graph $\mathcal{G}(m_1,m_2)$ is calculated as follows:

\begin{equation}
    \text{mean}(V_k, \mathcal{G}) = \frac{1}{|\mathcal{N}(V_k, \mathcal{G})|} \sum_{V_i \in \mathcal{N}(V_k, \mathcal{G})} E_i^{m_1}
\end{equation}

where \(\mathcal{N}(V_k, \mathcal{G})\) denotes the set of neighboring nodes for $V_k$, $V_i$ denotes the $i$-th node in $\mathcal{G}(m_1,m_2)$ corresponding to the embedding vector $E_i^{m_1} \in X^{m_1}$. 

This aggregate neighborhood information is then combined with the original features $E_k^{m_1}$ corresponding to $V_k$ in a concatenation operation, forming a comprehensive feature representation for further processing.

Following the aggregation step, the combined features undergo a linear transformation utilizing learned weights $W \in \mathbb{R}^{1024 \times C}$ and biases $b$. Here, $C=512$ denotes the number of output channels in the GraphSageConv operation.
 This transformation enables the model to learn intricate relationships within the graph structure. Furthermore, to maintain consistency and control the scale of the output, the features are subjected to L2 normalization. Mathematically, the GraphSageConv operation $\mathcal{Y}$ applied on the $k$-th sample can be summarized as follows:

\begin{equation}
\label{eq:graph-sage-conv}
    \mathcal{X}(k,X^{m_1},A^{m_2}) = \text{concat}(\text{mean}(V_k,\mathcal{G}(m_1,m_2)), {E_k^{m_1}}) \cdot W + b
\end{equation}

\begin{equation}
    \mathcal{Y}(k,X^{m_1},A^{m_2}) = \frac{\mathcal{X}(k,X^{m_1},A^{m_2})}{{\| {\mathcal{X}(k,X^{m_1},A^{m_2})} \|_2}}
\end{equation}

\begin{figure}[h]
    \centering
    \includegraphics[width=7 cm]{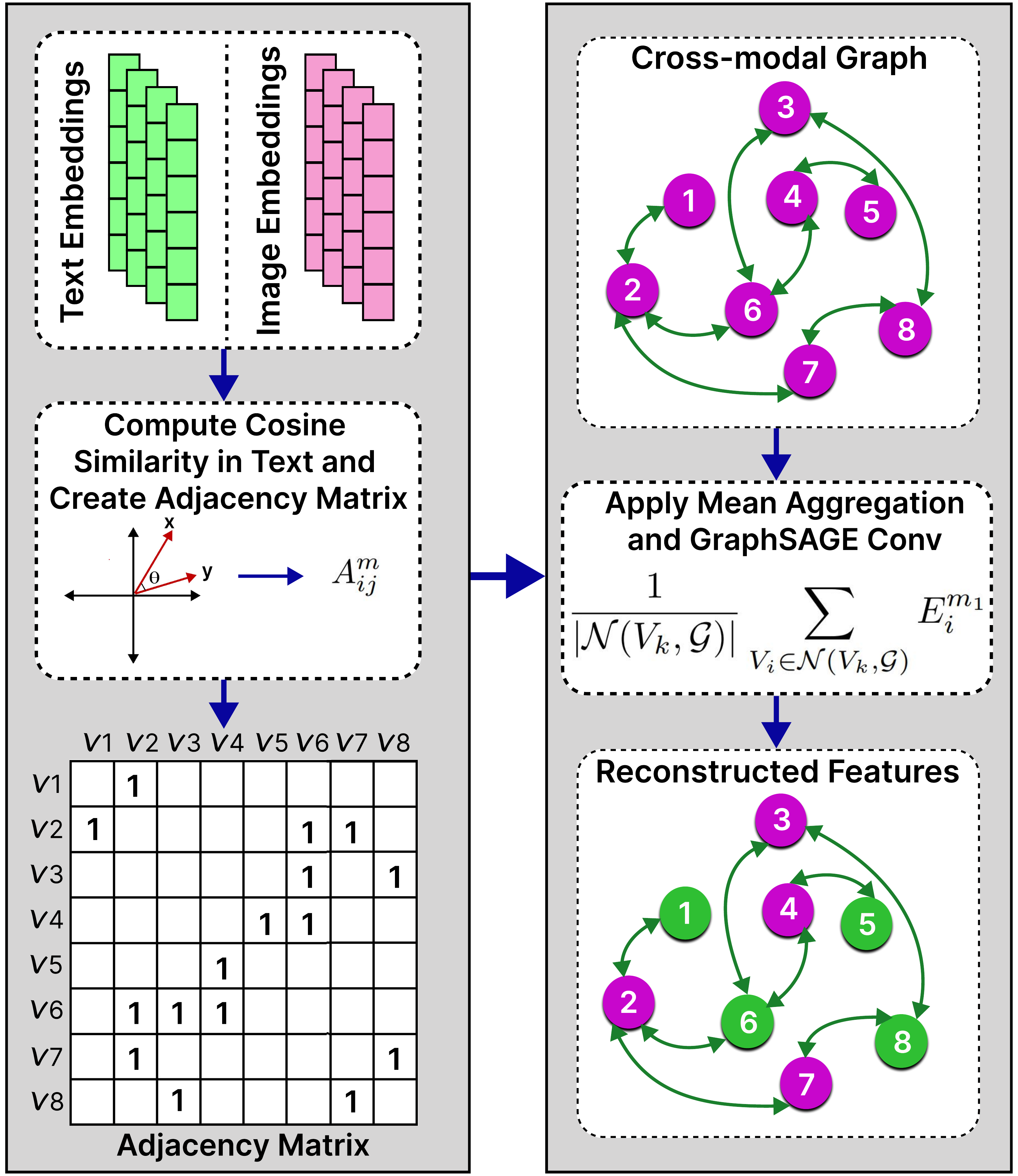}
    \caption{Cross-modal graph construction and updating: In the presented graph, vertices are image features, and edges between the vertices are constructed based on cosine similarities between the texts associated with the corresponding images. Another cross-modal graph is constructed with text features as vertices and image feature similarity as the criterion for edges.}
    \label{fig:cross-modal}
\end{figure}

We apply this operation to both the in-modal relation graphs, as represented in Equations \ref{eq:graph-II} and \ref{eq:graph-TT}, and both the cross-modal relation graphs, as represented in Equations \ref{eq:graph-IT} and \ref{eq:graph-TI}, constructed for a given meme. This allows us to obtain embedding vectors $H_k^{\text{img-img}}, H_k^{\text{txt-txt}}, H_k^{\text{img-txt}}, H_k^{\text{txt-img}}$. Finally, we concatenate all these vectors to form a cross-modal relation feature vector denoted as $H_k$, as shown in Equation \ref{eq:graph_concat}.

\begin{equation}
    \label{eq:graph-II}
    H_k^{\text{img-img}} = \mathcal{Y}(k,X^{\text{img}},A^{\text{img}})
\end{equation}
\begin{equation}
    \label{eq:graph-TT}
    H_k^{\text{txt-txt}} = \mathcal{Y}(k,X^{\text{txt}},A^{\text{txt}})
\end{equation}
\begin{equation}
    \label{eq:graph-IT}
    H_k^{\text{img-txt}} = \mathcal{Y}(k,X^{\text{img}},A^{\text{txt}})
\end{equation}
\begin{equation}
    \label{eq:graph-TI}
    H_k^{\text{txt-img}} = \mathcal{Y}(k,X^{\text{txt}},A^{\text{img}})
\end{equation}
\begin{equation}
    \label{eq:graph_concat}
    H_k = \text{concat}\left( H_k^{\text{img-img}}, H_k^{\text{txt-txt}}, H_k^{\text{img-txt}}, H_k^{\text{txt-img}} \right)
\end{equation}

\autoref{fig:in-modal} illustrates the construction and updating of the in-modal relation graphs, while \autoref{fig:cross-modal} depicts the construction and updating of the cross-modal relation graphs.

The proposed cross-modal relation learning utilizes sparse graphs, where edges are based on the cosine similarity threshold of other modality. In contrast, the traditional cross-attention-based fusion mechanism can be considered as a fully connected graph in a way that involves computing attention weights between every pair of nodes across different modalities. This means that each node attends to every other node, forming connections that can be represented as edges in a graph. The proposed CMRL method further uses the GraphSAGE algorithm to enhance feature representation, which adds to its run time complexity. However, the objective of CMRL is to reduce the impact of noise during the computation of multimodal features, trading off runtime complexity to enhance overall performance. The traditional cross-attention mechanism uses queries and keys of different modalities to calculate the similarity matrix using dot-product. This results in the noise of a modality influencing the multimodal representation. On the other hand, the proposed CMRL uses other modalities to build the edges between nodes. These edges, while representing the similarity between two modalities, do not directly interact with other modality's features. This way, the impact of noise is removed during the multimodal feature enhancement. The other simpler approach to reduce the impact of one 
 modality's noise could be the application of self-attention on individual modality features before concatenating them. However, in that way, the features of different modalities become independent of each other. We have shown in ablation analysis that the proposed approach outperforms both methods.

For Algorithm \ref{alg-graph}, graph construction (lines 2-6) has the most significant impact on the overall time complexity. In line 4, computing the similarity matrix \( S^m \) involves calculating the cosine similarity between pairs of \( D \)-dimensional embedding vectors for a total of \( N^2 \) pairs. The remaining steps can be disregarded as they do not significantly affect the overall time complexity. Therefore, the time complexity of Algorithm \ref{alg-graph} is \( O(N^2 \times D) \).

\subsection{Multifeature Learning and Classification}

This section describes a multifeature learner network that derives its inputs from the application of HIMA and CMRL, concatenated with task-specific features discussed in Section \ref{sec:task-specific-features}.
In the fusion process, we first concatenate the emotion feature vector, sentiment feature vector, and toxicity feature vector for the $k$-th sample to form a unified representation of task-specific features denoted by $T_k$. Then, we concatenate the embeddings obtained from the application of HIMA ($Z_k$), the embeddings derived from the application of CMRL ($H_k$), and $T$ to form a multifeature vector $M$ given by:

\begin{equation}
    M = \text{concat}(Z_k, H_k, T_k)
\end{equation}

Subsequently, $M$ is passed through a series of fully connected layers to obtain a hidden representation $H$, which serves as the input to multiple output layers, each corresponding to one of the tasks. In our case, since we are addressing five distinct tasks, we have five separate output layers. Each output layer is equipped with a softmax activation function, which normalizes the output scores across the different classes for its respective task. The output $\hat{y}_i$ of a task-specific output layer for task $i$ is computed as follows:

\begin{equation}
    \hat{y}_i = \text{softmax}(H \mathbf{W_i} + \mathbf{b_i})
\end{equation}

where $\mathbf{W_i}$ and $\mathbf{b_i}$ denote the trainable parameters corresponding to task $i$.

The loss function is defined using categorical cross-entropy for each task and summing it up. Let $T$ represent the number of tasks, $C_i$ denote the number of classes for task $i$, and $y_{ij}$ be the true label for class $j$ in task $i$. The loss function $\mathcal{L}$ is computed as:

\begin{equation}
\mathcal{L} = - \sum_{i=1}^{T} \sum_{j=1}^{C_i} y_{ij} \log(\hat{y}_{ij})
\end{equation}

where $\hat{y}_{ij}$ represents the predicted probability of class $j$ in task $i$. Minimizing this total loss during training ensures that the model learns to make accurate predictions across all tasks simultaneously.

\section{Experimental Evaluations}

\subsection{Datasets}
    \subsubsection{Memotion~\cite{sharma2020semeval}}
    The Memotion dataset, introduced as part of the SemEval-2020 Memotion Analysis task, is a comprehensive collection of 9,871 publicly available memes, each encompassing two key modalities: text and images. The dataset includes labels for sentiment, humor, sarcasm, offense, and motivation, making it particularly well-suited for tasks related to sentiment analysis, humor detection, sarcasm identification, offense detection, and motivational content recognition. Table \ref{tab:memotion_stats} represents the dataset statistics, including the number of original and augmented training samples, as well as the number of test samples for each category.

\begin{table}[htbp]
\centering
\caption{Memotion Dataset Statistics\label{tab:memotion_stats}}
\resizebox{\columnwidth}{!}{
\begin{tabular}{clcccccc}
\toprule
 Task & Label & \#Train\_orig & \#Train + Aug & \#Test \\
\midrule
\multirow{3}{*}{Sentiment} 
 & Negative & 629 & 4,948 & 173 \\
 & Neutral & 2,193 & 3,332 & 594 \\
 & Positive  & 4,138 & 6,923 & 1,111 \\
\midrule
\multirow{4}{*}{Humor} 
 & Not Funny & 1,646 & 4,297 & 445 \\
 & Funny & 2,442 & 3,782 & 654 \\
& Very Funny & 2,224 & 3,757 & 605 \\
 & Hilarious & 648 & 3,367 & 174 \\
\midrule
\multirow{4}{*}{Sarcasm} 
 & Not Sarcastic & 1,540 & 3,057 & 421 \\
& General & 3,490 & 5,688 & 937 \\
 & Twisted Meaning & 1,540 & 3,410 & 424 \\
 & Very Twisted & 390 & 3,048 & 96 \\
\midrule
\multirow{4}{*}{Offensiveness} 
 & Not Offensive & 2,699 & 4,337 & 707 \\
  & Slight & 2,583 & 4,662 & 709 \\
& Very Offensive & 1,459 & 3,770 & 387 \\
 & Hateful Offensive & 219 & 2,434 & 75 \\
\midrule
\multirow{2}{*}{Motivational} 
 & Not Motivational & 4,505 & 9,877 & 1,188 \\
& Motivation & 2,455 & 5,326 & 690 \\
\bottomrule
\end{tabular}}
\label{table:dataset_memotion}
\end{table}

\subsubsection{MMHS150K~\cite{Gomez_2020_WACV}} The MMHS150K dataset comprises 1,50,000 tweets with text and images collected from Twitter. It focuses on tweets containing 51 Hatebase terms associated with hate speech. Notably, the dataset includes both the raw, unfiltered content and associated images, providing a snapshot before content filters are applied. We use hate and not-hate classification of tweets. A total of 1,34,823 samples are utilized for training, 5,000 samples for validation, and 10,000 samples for testing. 

\subsubsection{HarMeme Dataset~\cite{pramanick2021detecting}}
The HarMeme dataset contains a total of 3544 memes. The dataset is labeled into three categories: very harmful, partially harmful, and harmless. For binary classification, we combine very harmful and partially harmful classes. In the final dataset, we get 1249 harmful and 2295 harmless memes. The dataset is split into training, validation, and testing with 3013, 177, and 354 samples, respectively.

\subsection{Baselines for Comparison}

 In this section, we provide an overview of the baseline methods utilized in our experiments, categorized into two groups: monomodal and multimodal techniques. Monomodal methods focus solely on either text or image information, whereas multimodal methods integrate both text and image data for analysis.

    BERT~\cite{devlin2018bert} and Mask RCNN-based detectron2 \cite{wu2019detectron2} monomodal methods are taken text and images, respectively. We utilize the ResNeXt-152-based (X152) Mask RCNN~\cite{he2017mask}model for image region features. 
    
    In multimodal methods we take Decision-level Fusion~\cite{gandhi2023multimodal}, Feature-level Fusion~\cite{gandhi2023multimodal}, MMCA~\cite{wei2020multi}, MMSA~\cite{vaswani2017attention}, CLIP~\cite{radford2021learning}, VisualBERT~\cite{li2019visualbert}, ViLT~\cite{kim2021vilt}, FCM~\cite{Gomez_2020_WACV}, DisMultiHate~\cite{3474085}, MSKAV+KDAC~\cite{CHHABRA2023106991}, Momenta~\cite{pramanick2021momenta}, MR.HA- RM~\cite{lin2023beneath}, MIMN~\cite{xu2019multi}, MIIM~\cite{pan-etal-2020-modeling}, MIMN-MTL~\cite{xu2019multi}, MIIM-MTL~\cite{pan-etal-2020-modeling}, HIMM~\cite{10068184}, Ensemsble~\cite{liu2024ensemble}, Ayetiran~\cite{ayetiran2024inter}, MMFHS~\cite{roy2025mmffhs}, ALFRED~\cite{sharma2024emotion}, and GPT-4o~\cite{key}. GPT-4o, the latest multimodal model from OpenAI, is employed in the prompt-based zero-shot classification approach.

\subsection{Results}

In this section, we compare the performance of our proposed method with the baseline methods mentioned in the previous section across three different datasets: Memotion, MMHS150K, and HarMeme. The comparison results are shown in \autoref{tab:memotion_results}, \autoref{tab:mmhs_results}, and \autoref{tab:harmeme_results}, respectively. It should be noted that all the scores are presented in percentages(\%).

\subsubsection{Performance Comparison on Memotion Dataset}

\begin{table*}[htbp]
\centering
\caption{Performance Comparison on Memotion Dataset\label{tab:memotion_results}}
\begin{tabular*}{\linewidth}{@{\extracolsep{\fill}} clcccccc }
\toprule
 & Method & Sentiment & Humor & Sarcasm & Offensive & Motivation \\
\midrule
\multirow{2}{*}{Monomodal } 
 & BERT~\cite{devlin2018bert} & 39.24 & 32.15 & 45.42 & 39.51  & 60.15\\
 & Mask RCNN X152~\cite{he2017mask} & 45.96 & 34.65 & 41.05 & 34.00 & 62.23\\
\midrule
\multirow{12}{*}{Multimodal} 
 & Decision-level Fusion~\cite{gandhi2023multimodal} & 46.19 & 36.81 & 45.65 & 40.77 & 62.66 \\
 & Feature-level Fusion~\cite{gandhi2023multimodal} & 45.86 & 36.43 & 46.06 & 40.82 & 62.51 \\
 & MMCA~\cite{wei2020multi} & 44.76 & 35.62 & 45.18 & 39.52 & 61.33 \\
 & MMSA~\cite{vaswani2017attention} & 46.30 & 36.97 & 46.11 & 41.34 & 62.89 \\
 & CLIP~\cite{radford2021learning} & 47.28 & 37.52 & 46.96 & 41.24  & 64.58\\
& VisualBert~\cite{li2019visualbert} & 44.26 & 34.47 & 43.74 & 39.85 & 62.44\\
 & ViLT~\cite{kim2021vilt} & 45.36 & 34.18 & 46.27 & 37.04 & 63.28\\
& MIMN~\cite{xu2019multi} & 35.34 & 29.90 & 47.18 & 35.34 & 59.61 \\
& MIIM~\cite{pan-etal-2020-modeling} & 36.12 & 31.26 & 45.05 & 34.37 & 50.29 \\
& MIMN-MTL~\cite{xu2019multi} & 42.91 & 34.56 & 48.54 & 35.15 & 63.11 \\
& MIIM-MTL~\cite{pan-etal-2020-modeling} & 40.97 & 35.15 & 48.35 & 36.31 & 63.30 \\
& HIMM~\cite{10068184} & 46.80 & 36.70 & 48.74 & 41.55 & 65.05 \\
 & Ensemble~\cite{liu2024ensemble} & 45.42 & 37.25 & 45.89 & 41.33 & 62.87\\
& GPT-4o~\cite{key}  &29.25 &34.12 &29.49 &35.42 &63.88 \\
\midrule
\multirow{1}{*}{} 
 & MM-ORIENT & \textbf{49.94} & \textbf{38.09} & \textbf{49.38} & \textbf{43.61} & \textbf{66.41}\\
\bottomrule
\end{tabular*}
\end{table*}

\autoref{tab:memotion_results} presents a comprehensive performance comparison of various methods on the Memotion dataset. To facilitate a meaningful comparison, we utilize the same number of samples as adopted by HIMM~\cite{10068184} for generating the results. Instances containing fewer than 10 words in the text are omitted. Following this step, the dataset is further partitioned into training, validation, and testing sets using an 8:1:1 ratio. However, for the ablation analysis, we utilize samples in the ratio depicted in \autoref{tab:memotion_stats}. To ensure consistency among the results, we use micro-F1 as adopted in MIMN, MIIM, MIMN-MTL, MIIM-MTL, and HIMM.

Notably, our proposed method stands out across all tasks. It achieves the highest micro-F1 scores in all the tasks. Monomodal methods, BERT (text) and Mask RCNN X152 (image) do not consider both modalities; hence, these models understandably provide lower performance than the proposed method. MM-ORIENT outperforms BERT by the highest margin of 10.70\% in the sentiment task and Mask RCNN X152 by 9.61\% in the offensive task. Feature-level fusion integrates both modalities before further processing, whereas decision-level fusion joins them after processing through the network. These methods do not apply any attention mechanism. MM-ORIENT outperforms these two methods in all the tasks, but the highest gain is achieved in the sentiment task, where the proposed method achieves 3.75\% and 4.08\% higher micro-F1 than decision-level fusion and feature-level fusion, respectively. The MMSA method, based on self-attention, outperforms the aforementioned fusion techniques, demonstrating the ability of monomodal attention to concentrate on discriminative information. Similarly, our proposed method employs monomodal attention, specifically HIMA, leading to enhancements in multitasking performance. The lower performance of MMCA denotes that cross-attention is influencing multimodal representation because of noise. Hence, the proposed approach learns multimodal representations without direct interaction between both modality features that reduces the influence of noise. The highest gain against MMCA and MMSA is achieved in the sentiment task, where the proposed method achieves 5.18\% and 3.64\% higher micro-F1, respectively.
Among the VisualBERT, ViLT, and CLIP methods, CLIP achieves competitive scores, indicating the contrastive learning-based approach can better understand and interpret multimodal content than the counterpart transformer-based models. Against CLIP, MM-ORIENT achieves the highest gain of 2.66\% in the sentiment task. Nevertheless, the intricate high-order intramodality and intertask relationships have not been thoroughly addressed in all these methods. GPT-4o, employed in zero-shot settings using the prompt-based approach, performs relatively less effectively than all the other methods. In the sentiment task, MM-ORIENT outperforms it by 20.69\% in terms of micro-F1. This indicates that large language models (LLMs) must be finetuned, as they may not perform well specific downstream tasks in zero-shot settings.

MIMN and MIIM methods, which have already been tested on the Memotion dataset, are analyzed. MIMN employs two memory networks for interaction between text and image representations, whereas MIIM employs a BERT-based approach and a co-attention module. Results indicate that these two methods could not comprehend the content effectively for multitasking. Multimodal-multitask methods, MIMN-MTL, MIIM-MTL, and HIMM, show significant performance improvement and generalization ability, with HIMM performing relatively better. Against HIMM, MM-ORIENT achieves a performance gain of 3.14\%, 1.61\%, 2.64\%, 1.06\%, and 1.36\% in sentiment, humor, sarcasm, offensive, and motivation tasks, respectively. These methods fall short of accurately capturing potential high-order relationships within and between modalities and overlook the facilitation among multiple tasks.
In comparison, the proposed method uses cross-modal relation learning using graphs to comprehend multimodal content and hierarchical interactive attention to facilitate capturing discriminative information to help in multitasking. The quantitative results demonstrate the efficacy of the proposed method in capturing high-order multimodal relationships and the generalizing ability for multitasking.

\subsubsection{Performance Comparison on MMHS150K Dataset}

\begin{table*}[htbp]
\centering
\caption{Performance Comparison on MMHS150K Dataset\label{tab:mmhs_results}}
\begin{tabular*}{\linewidth}{@{\extracolsep{\fill}} clcccccc }
\toprule
 & Method & Acc &P &R & F1 \\
\midrule
\multirow{2}{*}{Monomodal } 
 & BERT \cite{devlin2018bert}& 77.19 & 68.29 & 56.78 &62.01\\
 & Mask RCNN X152 \cite{he2017mask} & 70.16 & 56.35 & 55.30 &55.82 \\
\midrule
\multirow{10}{*}{Multimodal} 
 & Decision-level Fusion~\cite{gandhi2023multimodal} & 78.94 & 67.22 & 62.26 &64.64 \\
 & Feature-level Fusion~\cite{gandhi2023multimodal} & 78.73 & 66.74 & 62.45 &64.52 \\
 & MMCA~\cite{wei2020multi} & 77.61 & 65.24 & 60.85 &62.97 \\
 & MMSA~\cite{vaswani2017attention} & 79.17 & 67.58 & 63.05 &65.24 \\
 & CLIP~\cite{radford2021learning} & 79.26 & 68.83 & 63.57 &66.09 \\
& VisualBert~\cite{li2019visualbert} & 76.51 & 63.53 & 61.74 &62.62\\
 & ViLT~\cite{kim2021vilt} & 75.40 & 62.65 & 59.75 &61.17\\
 
& FCM~\cite{Gomez_2020_WACV} & 68.40 & - & - &70.40 \\
& DisMultiHate~\cite{3474085} & 76.54 & 67.51 & 71.42 &69.40 \\
& MSKAV+KDAC~\cite{CHHABRA2023106991} & 80.78 & 68.98 & 72.04 & 70.49 \\
 
 &
 Ayetiran~\cite{ayetiran2024inter} & 68.70 & 65.90 & 76.10 & 70.60\\
 & MMFHS~\cite{roy2025mmffhs} & 70.26 & 69.00 & 70.00 & 70.00\\
& GPT-4o~\cite{key}  &53.02 &51.68 &\textbf{93.27} &66.51 \\
\midrule
\multirow{1}{*}{} 
 & MM-ORIENT & \textbf{83.45} & \textbf{76.45} & 72.60 &\textbf{74.47}\\
\bottomrule
\end{tabular*}
\end{table*}

\autoref{tab:mmhs_results} provides a comprehensive performance comparison of different methods on the MMHS150K dataset across different metrics, including accuracy (Acc), precision (P), recall (R), and F1-score (F1). This dataset focuses on hate speech detection, a critical task in online content moderation.

Among the monomodal approaches, both BERT and Mask RCNN X152 are evaluated. BERT achieves an F1-score of 62.01\% and an accuracy of 77.19\%, demonstrating a solid understanding of textual content in identifying hate speech. However, Mask RCNN X152 lags slightly behind in accuracy at 70.16\%. This suggests that while image-based methods can contribute to hate speech detection, they may not be as effective as their text-based counterparts in isolation. MM-ORIENT outperforms BERT by a margin of 11.46\%  and Mask RCNN X152 by a margin of 18.65\% in terms of F1-score.

Results on the MMHS150K dataset denote that focusing on pertinent information before modality fusion helps in different tasks. This is affirmed by examining various fusion techniques, revealing that decision-level and MMSA fusion exhibit comparatively superior performance when compared to feature-level fusion and MMCA methods. This is attributed to the fact that both decision-level and MMSA methods prioritize individual modalities before combining them. The proposed MM-ORIENT achieves a performance gain of 11.5\% and 9.23\% against MMCA and MMSA, respectively, in terms of F1-score. CLIP outperforms transformer-based multimodal methods such as VisaulBERT and ViLT, suggesting that the multimodal contrastive learning approach has a superior ability to interpret multimodal content. These aforementioned techniques achieve satisfactory results but fall short in modeling complex intermodal and intramodal relationships. GPT-4o again acheives relatively lower scores than the other methods due to zero-shot settings.
Next, we analyze methods that have been proposed for the tasks specific to the MMHS150K dataset. FCM, DisMultiHate, MSKAV, Ayetiran and MMFHS methods also achieve competitive accuracy and F1-scores, with MSKAV performing relatively better than the others. MM-ORIENT outperforms MSKAV by a margin of 2.67\% and 3.98\% in terms of accuracy and F1-score, respectively.
The aforementioned three techniques generate high-order multimodal relationships via direct feature interaction at the latent stage, which influences pertinent monomodal information due to inherent noise. In contrast, our proposed approach generates cross-modally trained multimodal representation and avoids a direct interaction between monomodal features at the early stage. This is achieved by the cross-modal relation graphs. Additionally, proposed HIMA attention is effective in bringing out discriminative information from text and image features.

\subsubsection{Performance Comparison on HarMeme Dataset}

\begin{table*}[htbp]
\centering
\caption{Performance Comparison on HarMeme Dataset\label{tab:harmeme_results}}
\begin{tabular*}{\linewidth}{@{\extracolsep{\fill}} clcccccc }
\toprule
 & Method & Acc & P & R & Macro-F1 \\
\midrule
\multirow{2}{*}{Monomodal} 
 & BERT \cite{devlin2018bert} & 72.41 & 70.18 & 66.87 & 67.61 \\
 & Mask RCNN X152 \cite{he2017mask} & 66.95 & 64.01 & 64.06 & 64.04 \\
\midrule
\multirow{10}{*}{Multimodal} 
 & Decision-level Fusion~\cite{gandhi2023multimodal} & 73.85 & 71.51 & 71.58 & 71.54 \\
 & Feature-level Fusion~\cite{gandhi2023multimodal} & 72.41 & 70.31 & 71.19 & 70.61 \\
 & MMCA~\cite{wei2020multi} & 71.27 & 70.98 & 68.62 & 69.31 \\
 & MMSA~\cite{vaswani2017attention} & 75.00 & 72.73 & 72.11 & 72.38 \\
 & CLIP~\cite{radford2021learning} & 83.07 & 81.63 & 81.31 & 81.46 \\
 & VisualBert~\cite{li2019visualbert} & 79.27 & 78.93 & 77.80 & 78.46 \\
 & ViLT~\cite{kim2021vilt} & 74.85 & 72.90 & 74.33 & 73.32 \\
 & V-BERT COCO~\cite{li2019visualbert} & 81.36 & 79.55 & 81.19 & 80.13 \\
 & Momenta~\cite{pramanick2021momenta} & 83.82 & - & - & 82.80 \\
 & MR.HARM~\cite{lin2023beneath} & 86.16 & - & - & 85.43 \\
 
 & ALFRED~\cite{sharma2024emotion} & - & - & - & 85.88 \\
 & GPT-4o~\cite{key} & 67.51 & 63.55 & 58.27 & 57.62 \\
\midrule
 & MM-ORIENT & \textbf{88.32} & \textbf{87.74} & \textbf{86.37} & \textbf{86.84} \\
\bottomrule
\end{tabular*}
\end{table*}

\autoref{tab:harmeme_results} illustrates the performance of various methods on the Harmful Meme dataset, evaluated using metrics such as accuracy (Acc), precision (P), recall (R), and macro-F1 (F1). This dataset focuses on detecting harmful content in memes, an important aspect of content moderation.

For monomodal approaches, MM-ORIENT achieves a 19.23\% gain in macro-F1 compared to BERT, and a 22.80\% improvement over Mask RCNN X152. These results indicate that BERT performs slightly better with text-based content analysis compared to Mask RCNN X152 with image content. This indicates the challenges of using visual data alone for this task. Among the multimodal approaches, Decision-level Fusion and MMSA methods show competitive performance. CLIP, which employs a multimodal contrastive learning approach, significantly outperforms other methods but lags behind MM-ORIENT by a margin of 5.38\% in terms of macro-F1. VisualBert and ViLT also perform well, but not as effectively as CLIP.

Specialized methods for this dataset, such as Momenta, MR.HARM, and ALFRED achieve even higher performance, with MR.HARM reaching 86.16\% accuracy and 85.43\% F1-score. However, MM-ORIENT outperforms MR. HARM by 2.16\% and  1.41\% in terms of accuracy and macro-F1. This demonstrates the effectiveness of MM-ORIENT in leveraging both textual and visual data to accurately detect harmful memes.

\subsection{Ablation Analysis}

In this section, we present a detailed ablation analysis to gain a deeper understanding of the factors contributing to the performance of our proposed method. The analysis in this section focuses exclusively on the Memotion dataset to ensure a consistent and comprehensive examination of the model's performance. By isolating and modifying specific aspects of the model, we aim to shed light on the critical elements that enable the models to excel in multiple tasks.

Through a series of experiments, we systematically investigate the impact of different components and architectural choices on the model's performance on multiclass classification tasks (shown in Tables \ref{tab:abl-modalities}-\ref{tab:abl-additional}) as well as binary classification tasks (shown in Figures \ref{fig:abla_modality}-\ref{fig:abla_features}). Four tasks, humor recognition, sarcasm detection, offensive content identification, and motivation detection, are analyzed for binary classification, where the presence of that emotion is classified as one class, and the absence of that emotion is classified as another class. For instance, different humor levels, such as funny, very funny, and hilarious, are classified using a single label as humorous as they all depict the presence of humor, whereas not funny is classified with a different label as not humorous. The sentiment classification task is performed only in a multiclass setting. Micro-F1 is used as an evaluation metric for ablation analysis. 
    \subsubsection{Effect of Modalities}
\begin{figure}[htbp]
    \centering
    \includegraphics[width=7.88cm, height=4.8cm]{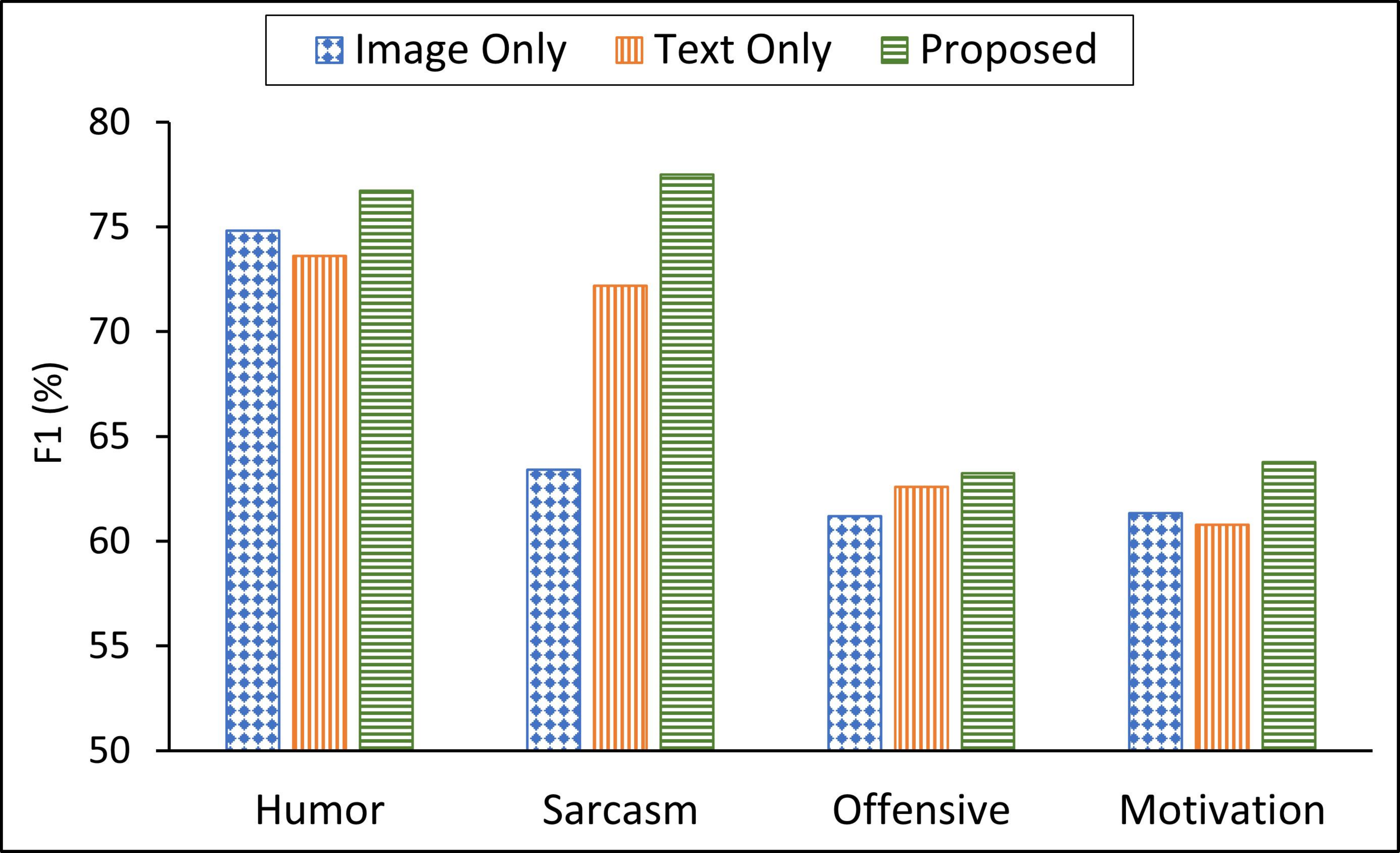}
    \caption{Influence of different modalities on binary classification performance}
    \label{fig:abla_modality}
    \end{figure}

    In order to assess the individual impact of each modality, we conduct experiments utilizing features exclusively from each modality. The results are shown in \autoref{tab:abl-modalities} and \autoref{fig:abla_modality} for multiclass and binary classification, respectively. 

    \begin{table}[htbp]
    \centering
    \caption{Ablation Analysis on Different Modalities\label{tab:abl-modalities}}
    \resizebox{\columnwidth}{!}{
    \begin{tabular}{lccccccccc}
    \toprule
     & Sentiment & Humour & Sarcasm & Offensive & Motivation \\
    \midrule
    Image Only & 54.52 & 31.81 & 36.59 & 35.16 & 61.34 \\
    Text Only & 53.01 & 29.13 & 45.47 & 37.21 & 60.79 \\
    MM-ORIENT & 58.20 & 35.49 & 49.50 & 39.45 & 63.78 \\
    \bottomrule
    \end{tabular}}
    \end{table}

    For the text modality, we pass the CLIP text features to the graph component, and BERT features to the attention component. The task-specific features are concatenated. Subsequently, the features from all these three components are concatenated and passed to fully connected dense layers. Similarly, for the image modality, we follow identical steps, using CLIP image features and Mask RCNN X152 features instead of the CLIP text features and BERT features. These are then directed to the image graph and region-based attention component, respectively. Given the absence of text in this analysis, toxicity BERT features are omitted from the additional features. The subsequent steps remain the same.
When excluding the text modality and exclusively utilizing the image modality, we observe a reduction in the micro-F1 score by 3.68\% for sentiment classification and by an average of 5.83\% and 5.12\% for multiclass and binary classification, respectively. It should be noted that for both multiclass and binary classification, the average is performed across all tasks mentioned in \autoref{tab:memotion_stats} except sentiment. Similarly, by removing the image modality and solely relying on the text modality, we see a decrease in the micro-F1 score by 5.19\% for sentiment classification and by an average of 3.91\% and 3\% for multiclass and binary classification, respectively. This notable drop in performance across all three tasks when the image modality is removed suggests the substantial significance of image modality compared to the text modality.
    \subsubsection{Effect of Attention Components} 
\begin{figure}[htbp]
    \centering
    \includegraphics[width=7.88cm, height=4.8cm]{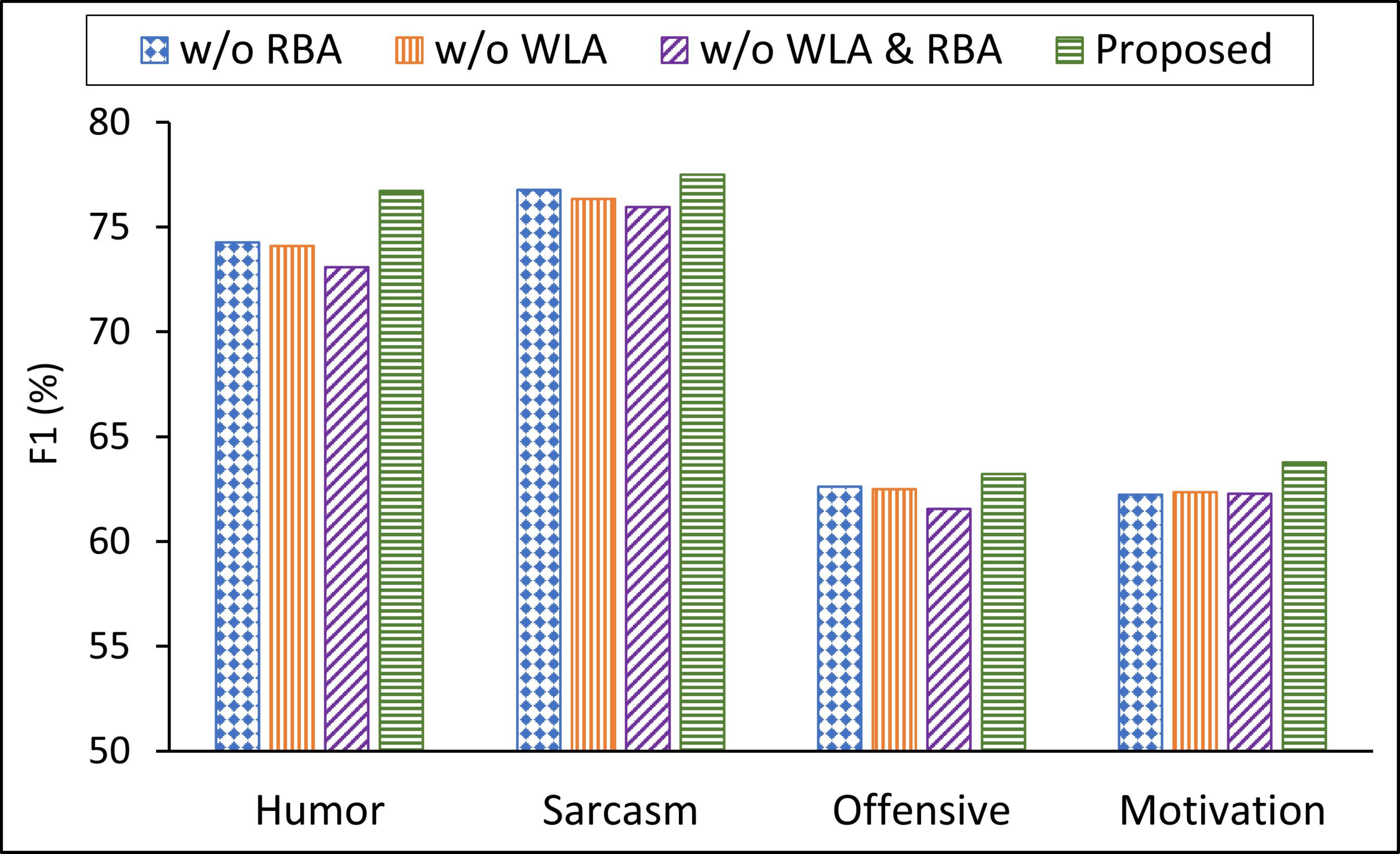}
    \caption{Influence of different attention mechanisms on binary classification performance}
    \label{fig:abla_attention}
    \end{figure}

    Investigating the attention components, we systematically examine the effects of both word-level attention and region-based attention. To do this, we conduct experiments where each attention component is removed while keeping the remainder of the architecture consistent. The results are summarized in \autoref{tab:abl-attention} and \autoref{fig:abla_attention} for multiclass and binary classification, respectively. 
    \begin{table}[htbp]
    \centering
    \caption{Ablation Analysis on Attention Components\label{tab:abl-attention}}
    \resizebox{\columnwidth}{!}{
    \begin{tabular}{lccccccccc}
    \toprule
     & Sentiment  & Humour & Sarcasm & Offensive & Motivation \\
    \midrule
    w/o RBA & 56.76  & 31.58 & 48.27 & 37.21 & 62.23 \\
    w/o WLA & 55.88  & 33.15 & 48.02 & 36.46 & 62.34 \\
    w/o RBA \& WLA & 54.72 & 30.13 & 47.30 & 36.04 & 61.28 \\
    MM-ORIENT & 58.20 & 35.49 & 49.50 & 39.45 & 63.78 \\
    \bottomrule
    \end{tabular}}
    \end{table}

    Firstly, when excluding region-based attention, we observe a notable reduction in the micro-F1 score across all tasks. Specifically, there is a decrease of 1.44\% for sentiment classification, 2.23\% for multiclass classification, and 1.34\% for binary classification. Similarly, omitting word-level attention leads to a decrease in performance, with a reduction of 2.32\% for sentiment classification, 2.06\% for multiclass classification, and 1.5\% for binary classification. Additionally, when both region-based and word-level attention are removed, we observe a substantial drop in performance, with a decrease of 3.48\% for sentiment classification, 2.82\% for multiclass classification, and 2.1\% for binary classification. These results underscore the critical role that both word-level and region-based attention play in enhancing the model's performance across a range of tasks.

    \subsubsection{Effect of Different Attention-based Fusion Mechanisms} 
\begin{figure}[htbp]
    \centering
    \includegraphics[width=7.88cm, height=4.8cm]{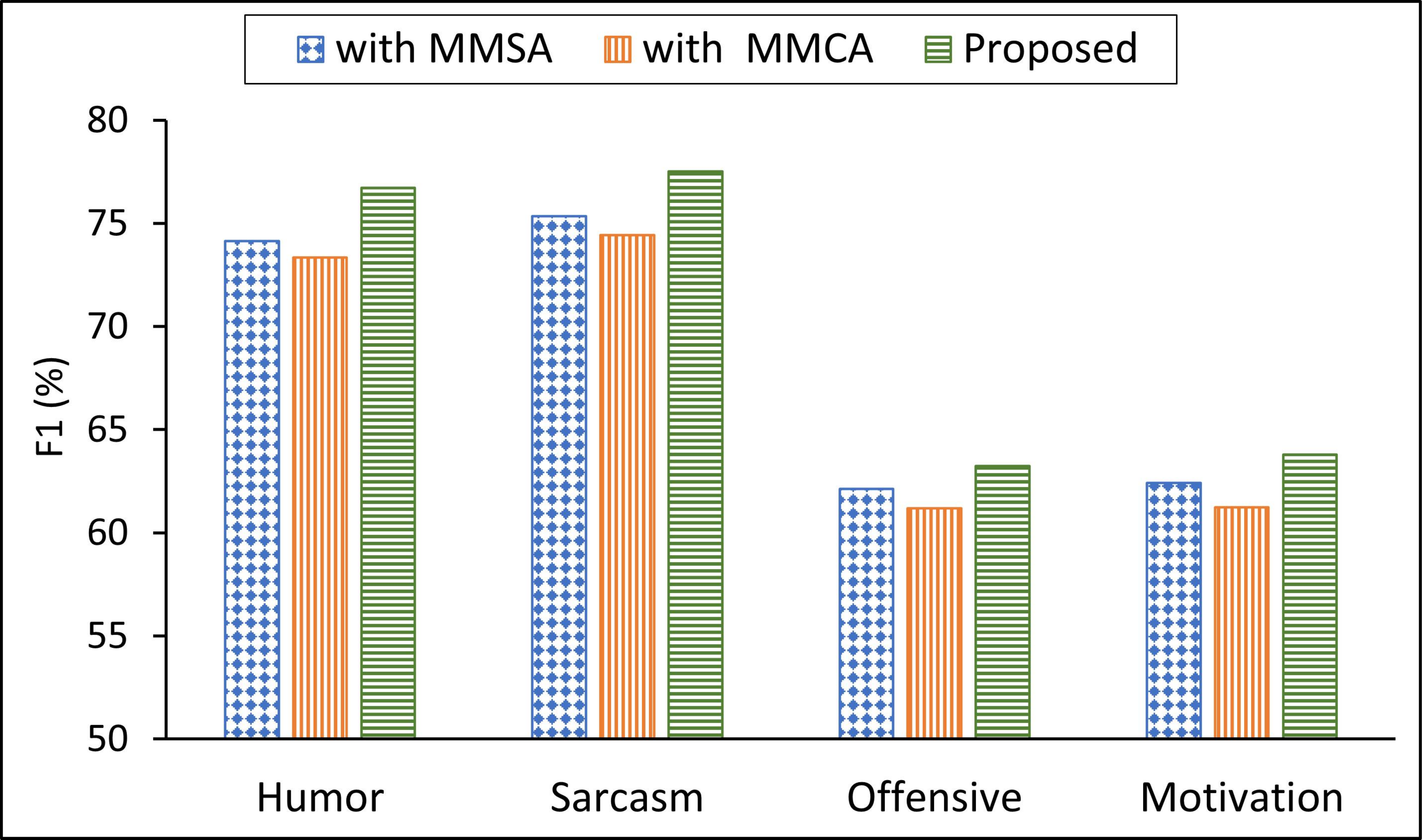}
    \caption{Influence of different attention-based fusion mechanisms on binary classification performance}
    \label{fig:abla_atten_Mech}
    \end{figure}

    To evaluate the impact of the proposed attention mechanism HIMA as compared to the different attention-based fusion mechanisms, we perform experiments using MMCA~\cite{wei2020multi} and MMSA~\cite{vaswani2017attention} while maintaining a consistent architecture. The outcomes are detailed in \autoref{tab:abl-atte_mech} and \autoref{fig:abla_atten_Mech} for multiclass and binary classification tasks, respectively. Using MMSA, we observe reasonable performance across all tasks in multiclass classification, with micro-F1 scores of 56.96\% for sentiment classification, 33.61\% for humor detection, 47.84\% for sarcasm detection, 37.43\% for offensive content detection, and 62.55\% for motivation classification. 

\begin{table}[htbp]
    \centering
    \caption{Effect of Different Attention-based Fusion Mechanisms}
    \label{tab:abl-atte_mech}
    \resizebox{\columnwidth}{!}{
    \begin{tabular}{lccccccccc}
    \toprule
    & Sentiment & Humour & Sarcasm & Offensive & Motivation \\
    \midrule
    with MMSA & 56.96 & 33.61 & 47.84 & 37.43 & 62.55 \\
    with MMCA & 56.17 & 32.71 & 46.68 & 37.26 & 61.35 \\
    MM-ORIENT & 58.20 & 35.49 & 49.50 & 39.45 & 63.78 \\
    \bottomrule
    \end{tabular}}
\end{table}

    MMCA mechanism results in slightly lower performance, particularly for sarcasm and motivation detection. In contrast, the MM-ORIENT mechanism consistently outperforms both MMSA and MMCA, achieving the highest scores across all tasks: 58.20\% for sentiment, 35.49\% for humor, 49.50\% for sarcasm, 39.45\% for offensive content, and 63.78\% for motivation. As shown in \autoref{fig:abla_atten_Mech}, in binary classification too, HIMA-based MM-ORIENT achieves higher scores across all tasks as compared to MMSA and MMCA. These findings underscore the efficacy of the HIMA method in improving MM-ORIENT's performance.

    \subsubsection{Effect of Graph Components} 
\begin{figure}[htbp]
    \centering
    \includegraphics[width=7.88cm, height=4.8cm]{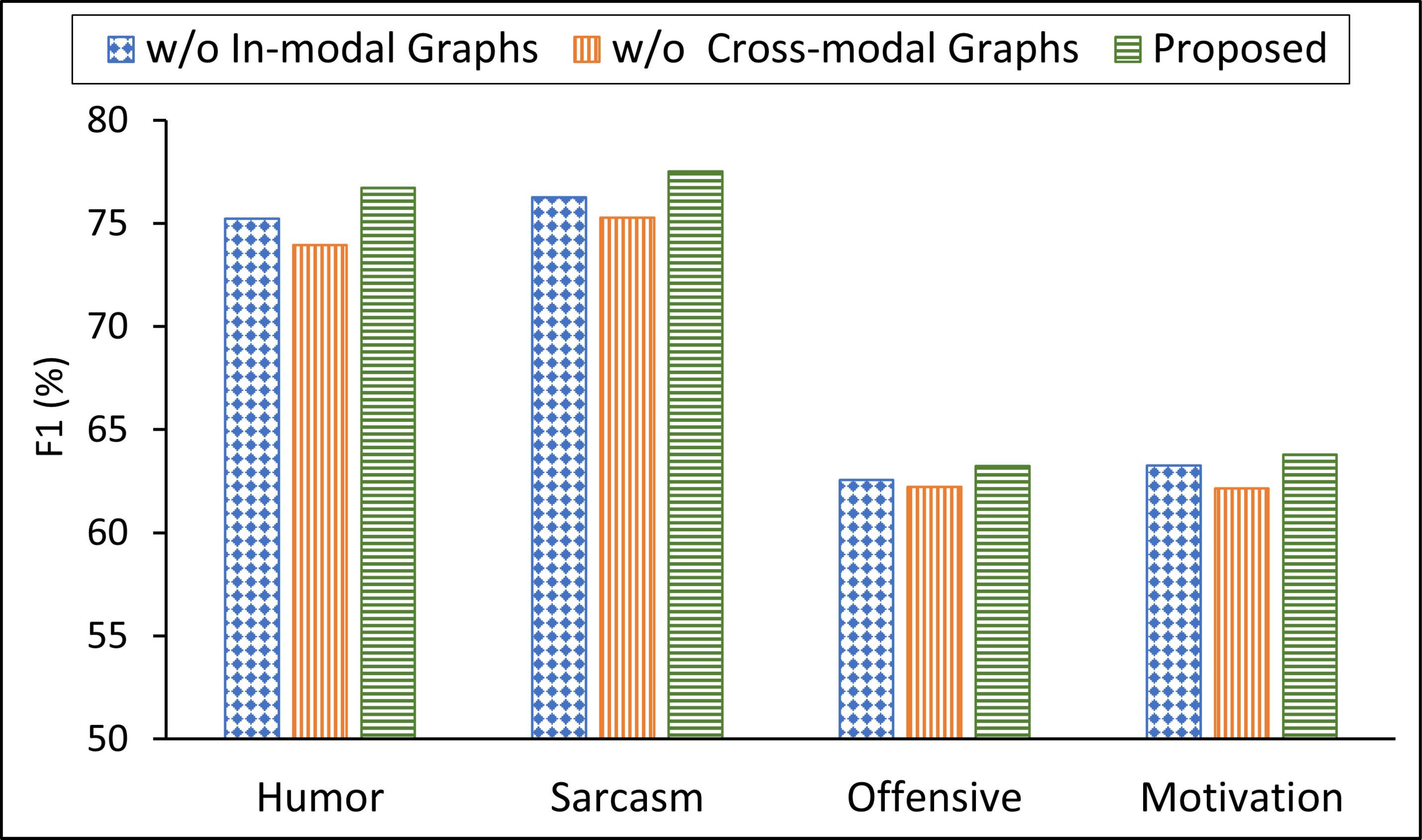}
    \caption{Influence of different graph mechanisms on binary classification performance}
    \label{fig:abla_graph}
    \end{figure}

    Examining the graph components, we systematically investigate the effects of both cross-modal and in-modal graphs. In order to do this, we conduct experiments where each graph component is removed while keeping the rest of the architecture unchanged. The results are summarized in \autoref{tab:abl-graph}. When excluding in-modal graphs, we observe a decrease of 1.93\% in micro-F1 for sentiment and an average of 1.62\%  decrease for other tasks in a multiclass setting. Moreover, as shown in \autoref{fig:abla_graph}, we observe a decrease of 1\% in the average micro-F1 score over four tasks in a binary classification setting. 
    \begin{table}[htbp]
    \centering
    \caption{Ablation Analysis on Graph Components \label{tab:abl-graph}}
    \resizebox{\columnwidth}{!}{
    \begin{tabular}{lccccccccc}
    \toprule
     & Sentiment  & Humour & Sarcasm & Offensive & Motivational \\
    \midrule
    w/o In-modal Graphs & 56.27  & 34.37 & 47.39 & 36.72 & 63.25 \\
    w/o Cross-modal Graphs & 55.72  & 32.15 & 46.30 & 35.04 & 62.28 \\
    MM-ORIENT & 58.20 & 35.49 & 49.50 & 39.45 & 63.78 \\
    \bottomrule
    \end{tabular}}
    \end{table}

    Results suggest that in-modal graphs help focus on intramodal feature refinement. However, without cross-modal relation graphs, there is a more pronounced impact on performance, suggesting that cross-modal graphs generate rich multimodal features. The micro-F1 score sees a substantial decrease of 2.48\% for sentiment and an average of 2.84\%  decrease for other tasks in a multiclass setting. For binary classification, a decrease of 1.92\% is observed in the average micro-F1 score over four tasks as shown in \autoref{fig:abla_graph}.
    These findings emphasize that cross-modal relation graphs provide efficient multimodal representations which in turn improve the model's performance. This confirms the ability of cross-modal relation graphs for effective semantic comprehension of image and text-based multimodal content.

    \subsubsection{Effect of Additional features} 
\begin{figure}[htbp]
    \centering
    \includegraphics[width=7.88cm, height=4.8cm]{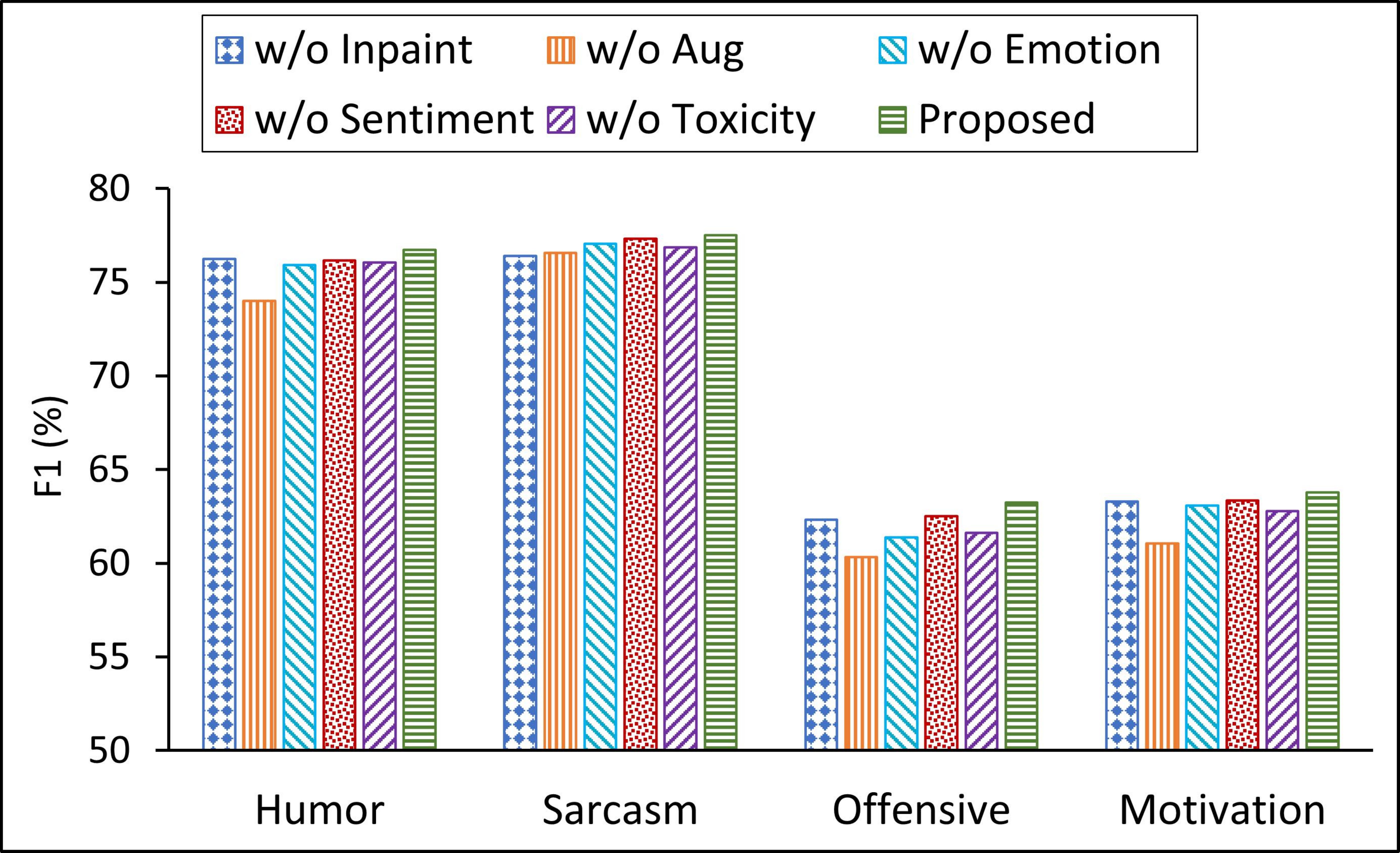}
    \caption{Influence of additional features on binary classification performance}
    \label{fig:abla_features}
    \end{figure}

    We systematically remove each of the features individually while keeping the rest of the architecture unchanged in order to evaluate the impact of each of these features as shown in \autoref{tab:abl-additional} and \autoref{fig:abla_features} for multiclass and binary classification respectively.

    \begin{table}[htbp]
    \centering
    \caption{Ablation Analysis on Additional Features\label{tab:abl-additional}}
    \resizebox{\columnwidth}{!}{
    \begin{tabular}{lccccccccc}
    \toprule
     & Sentiment & Humour & Sarcasm & Offensive & Motivation \\
    \midrule
    w/o Image Cleaning & 57.15  & 34.75 & 48.48 & 38.26 & 63.28 \\
    w/o Augmentation & 54.39  & 33.64 & 47.34 & 36.60 & 61.06 \\
    w/o Emotion & 56.27 & 34.37 & 47.75 & 37.72 & 63.07 \\
    w/o Sentiment & 57.89  & 35.12 & 48.88 & 39.26 & 63.34 \\
    w/o Toxicity & 57.48 & 34.24 & 48.24 & 38.33 & 62.79 \\
    MM-ORIENT & 58.20 & 35.49 & 49.50 & 39.45 & 63.78 \\
    \bottomrule
    \end{tabular}}
    \end{table}

    Firstly, without conducting image cleaning, we observe a decrease of 1.05\% for sentiment, an average of 0.86\% for multiclass classification, and an average of 0.75\% for binary classification in the micro-F1. Without the augmentation, we witness a substantial decrease in performance. The micro-F1 score drops by 3.81\% for sentiment, an average of 2.4\% for multiclass classification, and an average of 2.32\% for binary classification. Next, by excluding emotion features, observe a notable decrease in performance. The micro-F1 score diminishes by 1.93\% for sentiment, an average of 1.32\% for multiclass classification, and an average of 0.95\% for binary classification. Subsequently, removing the sentiment features, results in a marginal reduction of 0.31\% for sentiment, an average of 0.4\% for multiclass, and an average of 0.47\% for binary classification. Similarly, the removal of toxicity-related features results in a decrement of 0.72\% for sentiment, an average of 1.15\% for multiclass, and an average of 0.98\% for binary classification. These findings highlight the importance of each component in enhancing the model's performance across various tasks. The results denote that augmentation contributes the most to improve the model's performance as compared to other features as shown in \autoref{tab:abl-additional} and \autoref{fig:abla_features}.

\subsection{Qualitative Analysis}
    \begin{figure*}[t!]
    \centering
    \includegraphics[width=16cm, height=14cm]{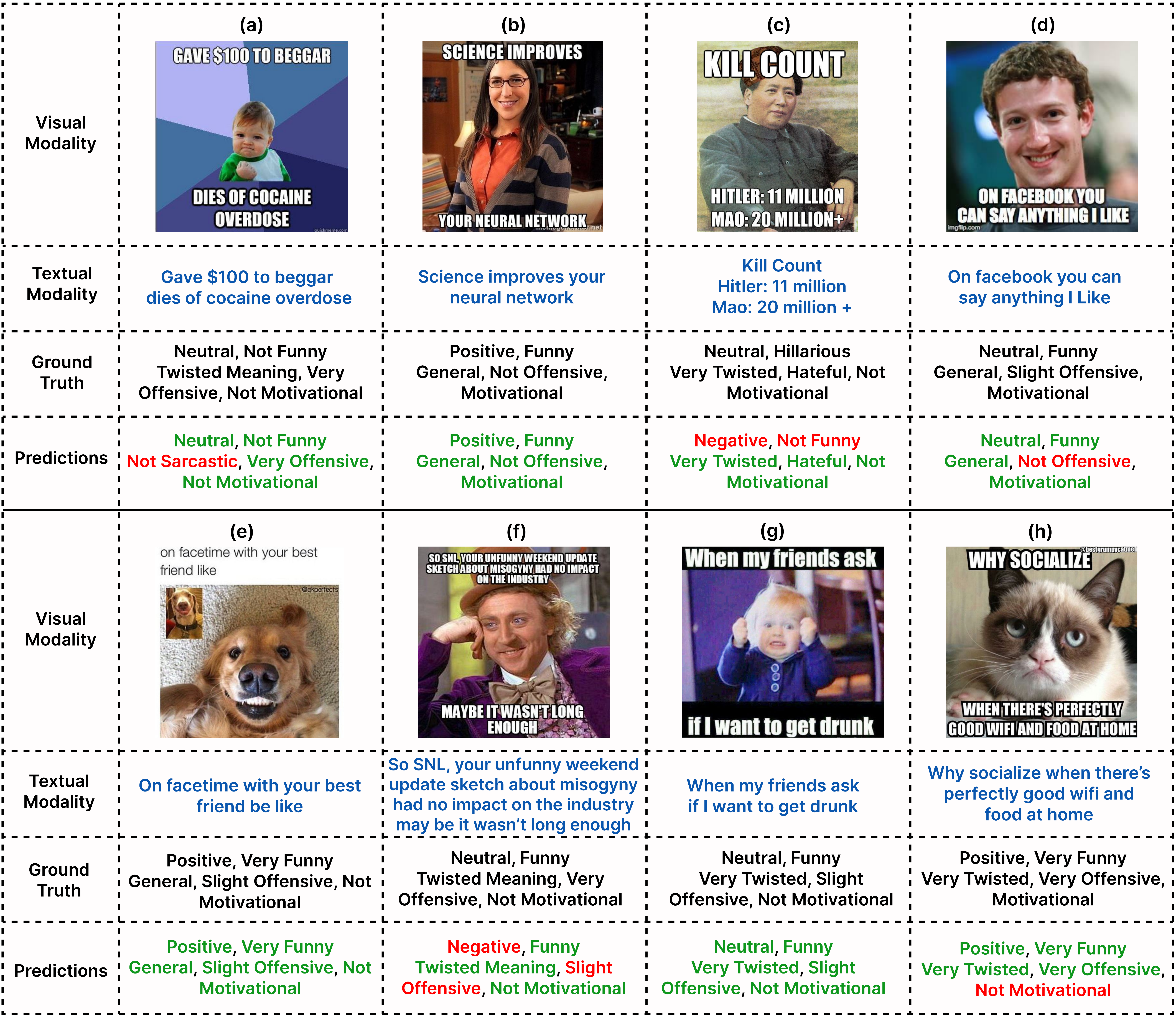}
    \caption{Predictions of MM-ORIENT on eight random samples from the Memotion dataset across tasks specified in \autoref{tab:memotion_stats}. First row: meme images, second row: text extracted from these images, third row: ground truth labels of the samples (sentiment, humor, sarcasm, offensiveness, motivation), fourth row: predictions of the model. The green color represents correct predictions, while the red ones are wrong predictions.}
    \label{fig:qualitative}
    \end{figure*}

The qualitative analysis of our meme comprehension framework involves a detailed examination of how the model interprets and categorizes different memes based on sentiment, humor, sarcasm, offensiveness, and motivation.

In Figure 9 (a), the visual component portrays a child with a determined facial expression, which, on its own, might convey a positive sentiment. However, a closer examination of the accompanying text, ``Gave \$100 to beggar dies of cocaine overdose," reveals a stark contrast, describing a negative outcome following what initially seems like a charitable act. This contradiction results in an overall neutral sentiment for the meme. Despite the ostensibly positive visual element, the meme, with its morbid twist, lacks humor and motivation. The model successfully predicts a neutral sentiment, appropriately recognizing the offensiveness of the content. However, despite the ironic interplay between the visual and textual components, the model falls short of capturing the sarcasm present in this particular instance.

Moving to Figure 9 (b), we encounter an image featuring a smiling woman wearing spectacles with the accompanying text, ``science improves your neural network." This combination delivers an overall positive and motivational message. There is an absence of any offensive content in both the visual and textual components, rendering the meme categorically ``Not Offensive." MM-ORIENT correctly predicts all the tasks for this meme.

In Figure 9 (c) features an image of Mao Zedong. The accompanying text provides a ``Kill Count" comparison, stating that Hitler is attributed to 11 million deaths, while Mao's count is indicated to be 20 million or more. Although the given label for sentiment is neutral, the correct sentiment of this meme is likely negative, as predicted by our model, due to the serious and grim nature of the content. Additionally, despite the actual label for humor being ``Hilarious," the meme doesn't appear to be designed for humor. The subject matter is sensitive, which is accurately captured by our model by classifying the meme as ``Not Funny." The seemingly straightforward presentation of death tolls creates a sarcastic contrast with the gravity of the subject matter. Consequently, the meme is appropriately categorized as ``Very Twisted" for sarcasm. The meme could be considered offensive, particularly to those with strong opinions about historical events and figures, warranting the highest degree of offensiveness classification - ``Hateful." 

The meme in Figure 9(d) is interpreted as negative due to its implication of restricted freedom of expression, despite being labeled as "funny" for its ironic humor. The meme sarcastically suggests that Facebook users can only express opinions that align with Mark Zuckerberg's preferences, creating a satirical tone. While it addresses content moderation on social media, the model incorrectly predicts it as "Not Offensive." Both our model and the actual label classify the meme as "Motivational," but a more accurate label would be "Not Motivational" due to its emphasis on restriction rather than inspiration. Mislabeling like this, though rare, can still occur in the dataset.

Similarly, the memes depicted in Figures 9 (e) through 9 (h) can be analyzed. In Figures 9 (e) and 9 (g), if we look at the text modality alone, it seems that the label for offensiveness should be ``Not Offensive," but then if we combine it with the visual modality, we can say that the label should be ``Slight Offensive" as given in the labels and our model's predictions. For each task, our model predicted the labels correctly in both of these figures. On the contrary, in Figure 9 (f), the text modality alone is sufficient to determine that the meme should be classified in the ``Very Offensive" category on the basis of offensiveness. It can be observed that our model falls slightly behind in this task and classifies it as ``Slight Offensive" for this sample. Interestingly, for the sentiment classification, despite the actual label being ``Neutral," our model predicts the label to be ``Negative," which seems more appropriate for this sample. Similarly, for Figure 9 (h), the model predicts the class as ``Not Motivational" for motivation detection, which seems to be the case here despite the actual label being ``Motivational." Hence, for figures from 9 (e) to 9 (h), our model predicts the relevant and appropriate labels across all tasks.

\section{Conclusion}
In this paper, we propose a novel multimodal-multitask framework, MM-ORIENT,  to comprehend visual and textual element-based multimodal content for multiple tasks such as sentiment, humor, offensiveness, and sarcasm detection. Our findings suggest that the inherent noise in monomodal features hampers the effectiveness of fusion techniques in creating efficient multimodal representations due to the direct interaction of features at the latent stage.
To address this, we propose a cross-modal relation graph-based approach for learning multimodal representations. This approach provides multimodal features without requiring direct interaction between features of different modalities in the initial stages. Our findings show that MM-ORIENT effectively reduces noise and generates high-level multimodal features in a cross-modal manner. The proposed attention mechanism, HIMA, employs a two-stage learning process to extract discriminative information from individual modalities. Additionally, our results indicate that generative AI-based augmentation enhances MM-ORIENT's performance as augmented samples help generalize the results. We have performed extensive experiments on three benchmark multimodal datasets to analyze the performance of the proposed method. In conclusion, the findings indicate that the proposed framework effectively cognizes multiple tasks in image and text-based multimodal content. By employing both quantitative and qualitative analyses on the Memotion dataset, we have shown that our method consistently surpasses the performance of existing advanced approaches.

\printcredits \\ \\

\balance
\bibliographystyle{unsrtnat}

\bibliography{cas-refs}
\balance

\end{document}